\documentclass[letterpaper]{article} % DO NOT CHANGE THIS
\usepackage{aaai25}  % DO NOT CHANGE THIS
\usepackage{times}  % DO NOT CHANGE THIS
\usepackage{helvet}  % DO NOT CHANGE THIS
\usepackage{courier}  % DO NOT CHANGE THIS
\usepackage[hyphens]{url}  % DO NOT CHANGE THIS
\usepackage{graphicx} % DO NOT CHANGE THIS
\usepackage{natbib}  % DO NOT CHANGE THIS AND DO NOT ADD ANY OPTIONS TO IT
\usepackage{caption} % DO NOT CHANGE THIS AND DO NOT ADD ANY OPTIONS TO IT
\usepackage{algorithm}
\usepackage{algorithmic}

\usepackage{booktabs}
\usepackage{graphicx}

\newcommand{\ie}{\emph{i.e., }}
\newcommand{\eg}{\emph{e.g., }}

\usepackage{subfigure}
\usepackage{subcaption}

\usepackage{array} 
\usepackage{amssymb} 
\usepackage{amsmath}
\usepackage{booktabs}
\usepackage{colortbl}
\usepackage[table]{xcolor}
\usepackage{makecell}
\usepackage{pifont}

\newcommand{\cmark}{\ding{51}} 
\newcommand{\xmark}{\ding{55}}

\usepackage{tabularx}

\usepackage{newfloat}
\usepackage{listings}
\DeclareCaptionStyle{ruled}{labelfont=normalfont,labelsep=colon,strut=off} % DO NOT CHANGE THIS
\floatstyle{ruled}
\newfloat{listing}{tb}{lst}{}
\floatname{listing}{Listing}
\title{Defeasible Visual Entailment: Benchmark, Evaluator, and Reward-Driven Optimization}
\author {
    Yue Zhang,
    Liqiang Jing,
    Vibhav Gogate
}
\affiliations {
    The University of Texas at Dallas\\
    \{yue.zhang,vibhav.gogate\}@utdallas.edu, jingliqiang6@gmail.com
}
\begin{document}

\maketitle

% Uncomment the following to link to your code, datasets, an extended version or similar.
%
% \begin{links}
%     \link{Code}{https://aaai.org/example/code}
%     \link{Datasets}{https://aaai.org/example/datasets}
%     \link{Extended version}{https://aaai.org/example/extended-version}
% \end{links}

\begin{abstract}

We introduce a new task called Defeasible Visual Entailment (DVE), where the goal is to allow the modification of the entailment relationship between an image premise and a text hypothesis based on an additional update. While this concept is well-established in Natural Language Inference, it remains unexplored in visual entailment. At a high level, DVE enables models to refine their initial interpretations, leading to improved accuracy and reliability in various applications such as detecting misleading information in images, enhancing visual question answering, and refining decision-making processes in autonomous systems. Existing metrics do not adequately capture the change in the entailment relationship brought by updates. To address this, we propose a novel inference-aware evaluator designed to capture changes in entailment strength induced by updates, using pairwise contrastive learning and categorical information learning. Additionally, we introduce a reward-driven update optimization method to further enhance the quality of updates generated by multimodal models. Experimental results demonstrate the effectiveness of our proposed evaluator and optimization method.
\end{abstract}

\section{Introduction}
Natural Language Inference (NLI) \cite{DBLP:conf/naacl/BosM05, DBLP:conf/mlcw/DaganGM05, DBLP:conf/iwcs/MacCartneyM09, DBLP:conf/emnlp/BowmanAPM15} is a fundamental task that involves determining the logical relationship between two sentences, specifically identifying whether one sentence entails, contradicts, or is neutral with respect to the other. 
% It utilizes the core concepts of entailment and contradiction \cite{kempson1977semantic, van2008brief}, essential for interpreting language meaning from words to texts. 
To further investigate the logical relationship across modalities, researchers have introduced a new inference task called Visual Entailment (VE). In VE, the premise is provided by an image and the hypothesis by a sentence and the task is to determine whether the image supports, contradicts, or is unrelated to the statement in the sentence \cite{DBLP:journals/corr/abs-1901-06706}.

%\textemdash consisting of image-sentence pairs
%whereby a premise is defined by an image, rather than a
%natural language sentence as in traditional NLI tasks \cite{DBLP:journals/corr/abs-1901-06706}. 
% the Stanford Natural Language Inference Visual Entailment (SNLI-VE) dataset \cite{DBLP:journals/corr/abs-1901-06706}
% Recognizing the ability of NLI tasks to evaluate fine-grained reasoning in complex inference scenarios within visual intelligence, researchers have introduced the Stanford Natural Language Inference Visual Entailment (SNLI-VE) dataset \cite{DBLP:journals/corr/abs-1901-06706}. This dataset is based on the foundational Stanford Natural Language Inference (SNLI) dataset \cite{DBLP:conf/emnlp/BowmanAPM15}, which is grounded in real-world scenarios and provides robust generalization capabilities for NLI models. This task integrates more complex multimodal data into the traditional NLI task and pays attention to the assessment of vision-language models' inferential capabilities. 
Existing approaches to the VE task typically leverage pre-trained vision-language models, such as OFA \cite{DBLP:conf/icml/WangYMLBLMZZY22}, UNITER \cite{DBLP:conf/eccv/ChenLYK0G0020}, FGAIF \cite{cite4}, and CoCa \cite{DBLP:journals/tmlr/YuWVYSW22}. These models are designed to understand and reason across visual and textual modalities and have greatly improved our ability to accurately link and interpret images and text.

Despite progress in this area, existing works on VE have mostly focused on clear, definite relationships and have not fully considered the uncertainties that can affect how images and text relate to each other. These uncertainties stem from factors such as incomplete information, unseen elements, image complexity, ambiguity, varying interpretations, differing perspectives, context, and the inherent subjectivity in visual perception.

\begin{figure}[t]
    \centering
    \includegraphics[width=\linewidth]{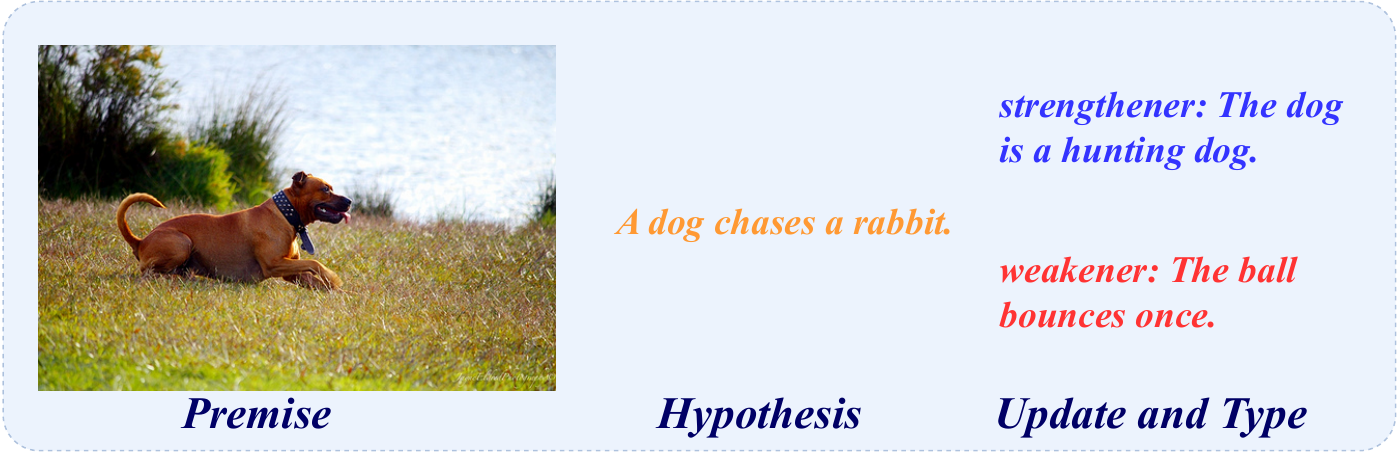}
    \caption{An example of defeasibility in visual entailment.}
    \label{fig:intro_case}
\end{figure}

To address this gap, we introduce the concept of \textit{Defeasible Visual Entailment} (DVE). The aim of DVE is to provide additional textual information that can either strengthen or weaken the relationship between an image premise and a text hypothesis. As illustrated in Figure \ref{fig:intro_case}, the premise shows a brown dog running through a grassy field. A strengthener could argue that ``The dog is a hunting dog,'' which strengthens the premise because a hunting dog is more likely to chase a rabbit. On the other hand, a weakener might state ``The ball bounces once,'' suggesting the dog is more likely chasing the ball than the rabbit.

%to weaken the likelihood that the dog is chasing a rabbit as the presence of the ball suggests the dog might be chasing the ball instead.

%One challenge with the DVE task is the lack of a benchmark specifically suited for it. 
A key challenge with the DVE task is that existing datasets used in visual entailment research are not suitable for evaluating and benchmarking methods designed to solve the DVE task. More specifically, previous benchmarks in visual entailment have primarily focused on definite relationships, often overlooking the role of defeasibility when uncertainties are present. Therefore, to fully harness the potential of defeasible inference in visual entailment, we introduce a new benchmark. To create this benchmark, we developed a new dataset for the DVE task by replacing the premises in the $\delta$-NLI dataset \cite{DBLP:conf/emnlp/RudingerSHBFBSC20} with images from the Flickr30k dataset \cite{DBLP:journals/tacl/YoungLHH14}. This approach minimizes costs while maximizing the use of existing resources. In our dataset, each premise-hypothesis pair includes multiple strengtheners and weakeners.

Building upon this dataset, we propose two specific DVE tasks: (1) \textit{Classification} Task: predicting whether a provided update (sentence) acts as a strengthener or a weakener for the premise-hypothesis pair. 
(2) \textit{Generation} Task: given a premise-hypothesis pair as input, generate an update sentence that weakens or strengthens the hypothesis. While the classification task can be easily evaluated using accuracy, the generation task lacks a metric that effectively captures the change in entailment strength introduced by the generated update. An ideal metric should reflect how the update influences the increase or decrease in entailment strength. 

%It is easy to evaluate the classification task by accuracy. But for the generation task, there is no metric that can capture the change of entailment strength brought by the generated update.
%An ideal metric should reflect the increase or decrease in entailment strength with the incorporation of updates. 

To address this issue, we propose a new \textit{evaluator} capable of measuring the change of entailment strength brought by the generated update. We also introduce a learning scheme that employs pairwise contrastive learning and categorical information learning to train the evaluator in an unsupervised manner. Our evaluator outputs a value representing the entailment strength for a given triplet (update, premise, hypothesis). We conducted a human evaluation to compare the performance of our evaluator with existing metrics, such as ROUGE-L \cite{lin2004rouge}, and CLIPScore \cite{DBLP:conf/emnlp/HesselHFBC21}. 
Our experimental results demonstrate that our metric achieves the best correlation with human evaluation results and existing metrics are unable to accurately capture the change of reasoning relationship brought by the update.

%In our experiments, we observed that the generated updates are not good enough, which usually can not change the entailment relationship of the premise-hypothesis pair. Therefore, we also devise a reward-driven update optimization, which can utilize the evaluation results of our evaluator to further refine the generated updates. The experimental results show our method can generate better updates than baselines.
%In conclusion, our contributions can be summarized as follows,
In our experiments, we found that directly adapting existing VE methods for the DVE task (baseline approaches) results in low-quality updates, which often fail to alter the entailment relationship between the premise and hypothesis. To address this, we developed a reward-driven update optimization technique that leverages the evaluation results from our evaluator to further refine the generated updates. Our experimental results demonstrate that this new method produces higher-quality updates compared to baseline approaches. 
In summary, our contributions are:
\begin{enumerate}
    \item We propose a defeasible visual entailment task and build the first benchmark\footnote{Our code and data are available at \url{https://github.com/skywalkerzhang/Defeasible_Visual_Entailment}.} for it. This benchmark enables a thorough investigation of the fine-grained multimodal understanding capabilities of state-of-the-art models. 
    %which can investigate the fine-grained multimodal understanding ability of multimodal models.
    % Defeasible Inference in the multimodal domain and proposed two specific tasks within this benchmark.
    \item We devise a novel inference-aware evaluator that leverages advanced pairwise contrastive learning and categorical information learning for capturing the change of entailment strength brought by the update.
    \item We propose a new reward-driven update optimization method and demonstrate experimentally that our method significantly enhances the quality of generated updates, outperforming state-of-the-art models. %can enhance the generation quality of the state-of-the-art model.
\end{enumerate}

\section{Task Defination}
In this paper, we follow the definition of the defeasible task \cite{DBLP:conf/emnlp/RudingerSHBFBSC20}
\begin{quote}
    Given an image premise $ I$, a hypothesis $ H $ is defeasible if there exists an update $ {U} $ (consistent with $ {I} $) such that a human would find $ {H} $ less likely to be true after learning $ {U} $. Specifically, an update $ {U} $ is called a weakener if given a premise $ {I} $ and hypothesis $ {H} $, a human would most likely find $ {H} $ less likely to be true after learning $ {U} $; if they find $ {H} $ more likely to be true, then we call $ {U} $ a strengthener.
\end{quote}
\subsection{Classification Task}
\subsubsection{Formulation}
\label{sec:cls-task}
% We introduce a dataset $\mathcal{D}_c$ for this task formulated as $\mathcal{D}=\{(I_1, H_1, U_1, G_1), (i_2, h_2, u_2, g_2), \ldots, (i_n, h_n, u_n, g_n)\}$, where $(i_k, h_k, u_k, g_k)$ is an instance from $\mathcal{D}$, with $i_i$, $h_k$, $u_k$, and $g_k$ denoting an image premise, a text hypothesis, an update, and a goal, respectively. 

The goal of the classification task is to find a classification model $\mathcal{M}_c$ which predicts the update type based on the premise $I$, hypothesis $H$, and update $U$ as follows,
\begin{equation}
    \hat{L} = \mathcal{M}_c (I, H, U),
\end{equation}
where $\hat{L} \in \{w, s\}$ denotes the predicted update type. 
% (\ie $l$ (strengthener) or $w$ (weakener)) of the update $U$ based on the relationship conveyed by $(i_k, h_k, u_k)$. 
% $s$ (strengthener) and $w$ (weakener), are assigned based on the relationship conveyed by $(i_k, h_k, u_k)$. 
% As mentioned before, 
$\hat{L} = s$ (strengthener) is assigned if $U$ makes the hypothesis $H$ more likely given the image $I$ while
$\hat{L} = w$ (weakener) is assigned if $U$ makes the hypothesis $H$ less likely given the image $I$.

\subsubsection{Evaluation Metric}
% To evaluate the performance of the model $\mathcal{M}_c$ on the classification task, 
We use accuracy as the  metric. 
% Accuracy measures the proportion of correctly classified instances among the total instances, providing a straightforward assessment of model performance. 
% Formally, accuracy is defined as:
% \begin{equation}
% \text{Accuracy} = \frac{TP + TN}{TP + TN + FP + FN},
% \end{equation}
% where $TP$ represents true positives, $TN$ represents true negatives, $FP$ represents false positives, and $FN$ represents false negatives.

\subsection{Generation Task}
\label{sec:gen-task}
\subsubsection{Formulation} 
% Similarly, we also use the defined dataset $\mathcal{D}$ for the generation task, defined as $\mathcal{D} = \{(I_1, H_1), (i_2, h_2), \ldots, (i_n, h_n)\}$, where $(i_k, h_k)$ is an $i$-th instance from $\mathcal{D}$, with $i_k$ and $h_k$ denoting an image premise and a text hypothesis, respectively.

In this task, the model aims to generate an update based on the input premise $I$, hypothesis $H$, and goal $G \in \{w,s\}$ (\ie weakener or strengthener) as follows,
\begin{equation}
    \hat{U} = \mathcal{M}_g (I,H,G),
\end{equation}
where $\hat{U}$ is the generated (textual) update.% from the generation model $\mathcal{M}_g$. $\hat{U}$ should follow the goal of the input $G$. 

% Specifically, the model should generate:
% i) $u_k$ such that $i_k \models (h_k \wedge u_k)$, if the goal is to strengthen the hypothesis $H$;
% ii) $u_k$ such that $i_k \models (h_k \wedge \neg u_k)$, if the goal is to weaken the hypothesis $h_k$.

\subsubsection{Evaluation Metric} 
To comprehensively assess the quality of the generation model $\mathcal{M}_g$, we utilize a variety of evaluation metrics,  including traditional evaluation metrics: ROUGE-L \cite{lin2004rouge}, BLEU-4 \cite{DBLP:conf/acl/PapineniRWZ02}, deep learning-based metrics: BERTScore \cite{DBLP:conf/iclr/ZhangKWWA20} and CLIPScore \cite{DBLP:conf/emnlp/HesselHFBC21}, and our custom-designed reference-free Inference-aware Evaluator, which is detailed in the later section.
\section{Defeasible Visual Entailment Dataset}
\subsection{Dataset Construction}

In this section, we describe the construction of our dataset for the DVE task, which leverages three existing datasets:  Flickr30k \cite{DBLP:journals/tacl/YoungLHH14}, SNLI \cite{DBLP:conf/emnlp/BowmanAPM15} and the $\delta$-NLI dataset \cite{DBLP:conf/emnlp/RudingerSHBFBSC20}. 

Flickr30k is a well-known image captioning dataset comprising 31,783 images and 158,915 captions, depicting everyday activities, events, and scenes. Each image in the dataset is annotated with five captions generated through crowdsourcing, providing diverse descriptions of the visual content. This dataset is essential for developing models that can understand and generate natural language descriptions of images, as it offers a rich set of image-caption pairs that cover a broad range of scenarios and objects.

The SNLI dataset is a large annotated textual entailment dataset. It comprises approximately 570,000 premise-hypothesis $(T, H)$ pairs, as well as their corresponding label categorized into three classes: entailment, neutral, and contradiction. 
The premise was originally collected from the captions in Flickr30k.
The hypothesis was written via Amazon Mechanical Turk for each class.  
% The neutral pairs, which are of particular interest in our work, are those where the hypothesis is neither entailed nor contradicted by the premise, allowing for potential strengthening or weakening of the statement under appropriate conditions. 
% Data validation in SNLI involves assigning a gold label to each pair based on the consensus of at least 3 out of 5 crowdsourcing workers.

The $\delta$-NLI dataset is designed to collect strengtheners and weakeners for the NLI task, which can be used to further investigate the semantic understanding ability in models.  The new dataset was devised for the defeasible inference tasks in natural language.
% demonstrating the generality of the defeasible inference framework. 
This dataset 
% includes a subset specifically for Natural Language Inference, named $\delta$-SNLI, which 
contains 10,000 neutral premise-hypothesis pairs derived from the SNLI dataset. 
In the context of SNLI, neutral premise-hypothesis pairs are those where the hypothesis is neither entailed nor contradicted by the premise, thereby making it easy to issue additional information to strengthen or weaken the statement under appropriate conditions. 
The premise is from the captions in the Flickr30k dataset. Crowdsourced workers were assigned the task of writing updates, including both strengtheners and weakeners.

\begin{figure}[t]
    \centering
    \includegraphics[width=\linewidth]{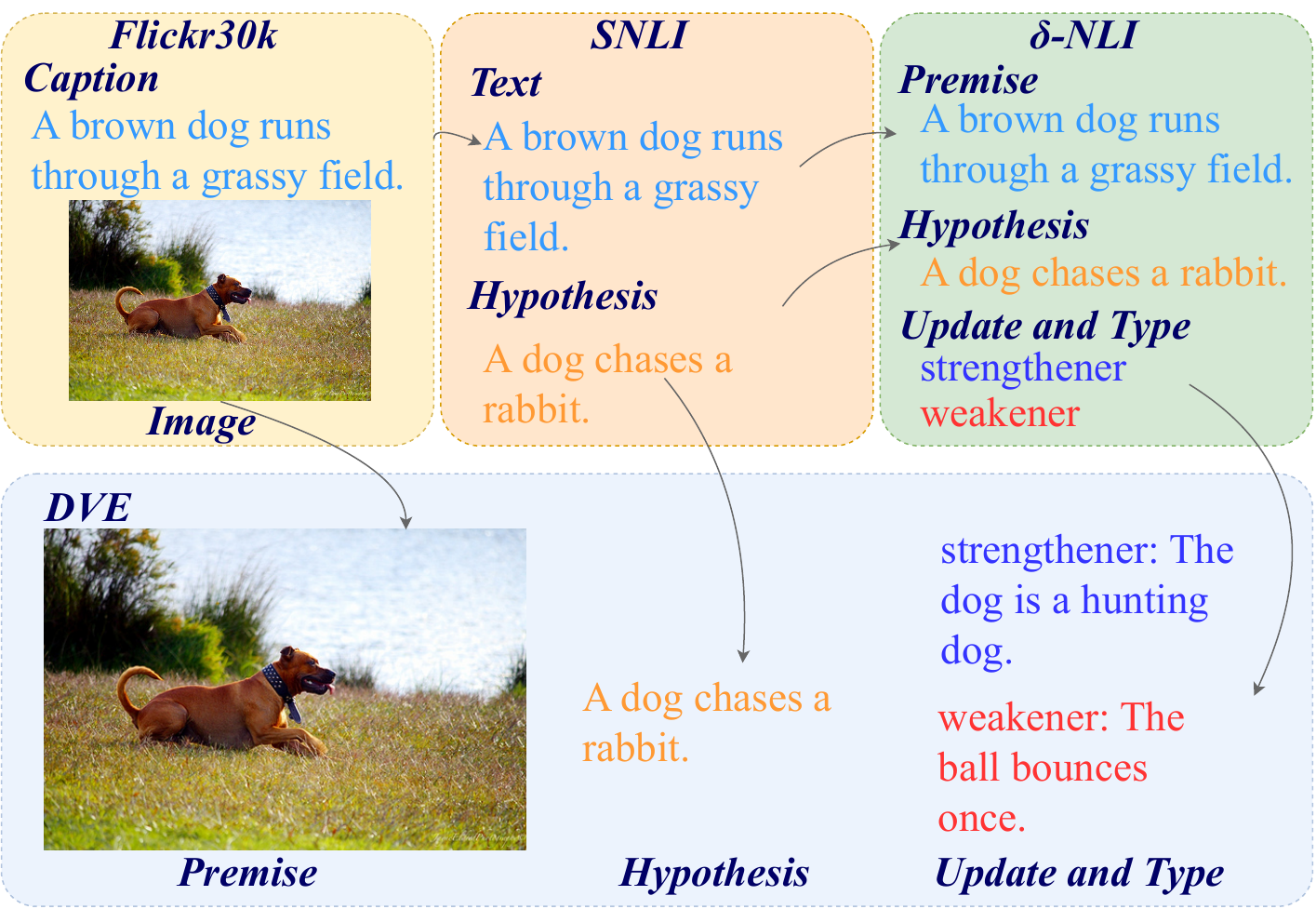}
    \caption{The workflow of generating the DVE dataset by integrating premises and hypotheses from SNLI with images from Flickr30k and updates from $\delta$-NLI.}
    \label{fig:dataset_generation}
\end{figure}

Although existing datasets have been successful in assessing the semantic entailment capability of models, defeasibility in the visual domain has not been explored. Therefore, our work focuses on creating a novel dataset for DVE, which consists of image premises, text hypotheses, and updates (including weakeners and strengtheners) for premise-hypothesis pairs.
To simplify and save cost, we constructed our new dataset based on the Flickr30k, SNLI, and $\delta$-NLI datasets. 
% Given that the defeasible inference dataset utilizes images from Flickr30k, it is feasible to integrate these datasets to construct the VDI dataset. 
Specifically, 
for each premise-hypothesis pair $(T, H)$ pair in the SNLI dataset, we replace the text premise with its corresponding image in Flickr30k, with the premise-hypothesis pair formulated as $(I, H)$. Thereafter, we incorporate the update from $\delta$-NLI into our DVE dataset. We only retain the premise-hypothesis pair that has an update in the $\delta$-NLI dataset. The overall workflow is shown in Figure \ref{fig:dataset_generation}.

% This results in a $(i_{\text{image}}, h_{\text{text}})$ pair in the MDI dataset, facilitating a structured dataset that supports the exploration of visual defeasible inference.

\begin{table}[t]
    \centering
    \resizebox{\columnwidth}{!}{
    \begin{tabular}{lccc}
        \toprule
        \textbf{Statistics} & \textbf{\makecell{Train \\ set}} & \textbf{\makecell{Validation \\ set}} & \textbf{\makecell{Test \\ set}} \\
        \midrule
        Total samples & 93,082 & 1,888 & 1,972 \\
        Update type dist. &  &  &  \\
        \quad Weakener & 46,541 & 944 & 986 \\
        \quad Strengthener & 46,541 & 944 & 986 \\
        Average premise length & 12.83 & 13.82 & 13.21 \\
        Average hypothesis length & 8.27 & 8.41 & 8.23 \\
        Unique premises & 9,293 & 191 & 200 \\
        Unique hypotheses & 9,438 & 195 & 203 \\
        Average updates per image & 9.79 & 9.68 & 9.71 \\
        Unique images & 9,507 & 195 & 203 \\
        \bottomrule
    \end{tabular}
    }
    \caption{Statistics of the DVE dataset.}
    \label{tab:dve_stats}
\end{table}

\subsection{Statistics of DVE}
In this section, we present the statistical overview of the DVE dataset, divided into training, development, and test sets. The statistics are summarized in Table \ref{tab:dve_stats}.
% The DVE dataset includes 93,082 samples in the training set, 1,888 in the development set, and 1,972 in the test set, ensuring ample data for robust model training and evaluation. Each subset maintains an equal distribution of weakeners and strengtheners, with 46,541 samples for each type in the training set, around 944 samples in the development set, and 986 in the test set.
% The average lengths of premises and hypotheses are consistent, with premises averaging 12.83 words in the training set, 13.82 in the development set, and 13.21 in the test set. Hypotheses average around 8 words across all sets. This consistency helps in maintaining uniform task complexity.
% The dataset is diverse, with over 9,000 unique premises and hypotheses in the training set. Each image has around 9-10 updates on average, providing a rich variety of data. Additionally, there are over 9,500 unique images, ensuring exposure to diverse visual contexts.
Overall, the DVE dataset's balanced and diverse data support comprehensive training and evaluation of models on visual defeasible inference tasks.
We further compare our DVE dataset with the related datasets in the supplementary material.

\section{Estimating the Impact of Updates on Multimodal Defeasible Reasoning}
\label{sec:custom_eval}

For the generation task, we utilize standard generation evaluation metrics such as BLEU, ROUGE, and BERTScore to measure the quality of the generated updates. 
These metrics assess the lexical or semantic similarity between the generated update and the reference updates, but it is not realistic to collect a comprehensive set of ground-truth references for such open-domain tasks, where answers can vary widely. We also employ the reference-free metric CLIPScore, which primarily evaluates the similarity between the answer and the image. 
% Therefore, it is challenging to devise a metric to accurately capture the changes in inference strength brought by strengtheners and weakeners. 
While these metrics provide some insight into the quality of the updates, they are not well-suited to accurately capture the changes in entailment strength brought about by strengtheners and weakeners.
% Traditional metrics often fail to reflect the nuanced impact of these updates. 
% An ideal metric should reflect the increase or decrease in inference strength with the incorporation of these updates.
To address this, we propose a new reference-free evaluation approach utilizing contrastive learning to train an unsupervised model capable of representing the entailment strength of the changes caused by updates. 
% Our method aims to learn a robust representation of the updates in relation to the premise and hypothesis, capturing the nuanced impact of strengtheners and weakeners. 
% This approach allows us to evaluate the generated updates based on their effect on the inference strength, providing a more context-sensitive measure of performance.

\subsection{Inference-aware Evaluator}
\begin{figure*}[t]
    \centering
    \includegraphics[width=0.95\linewidth]{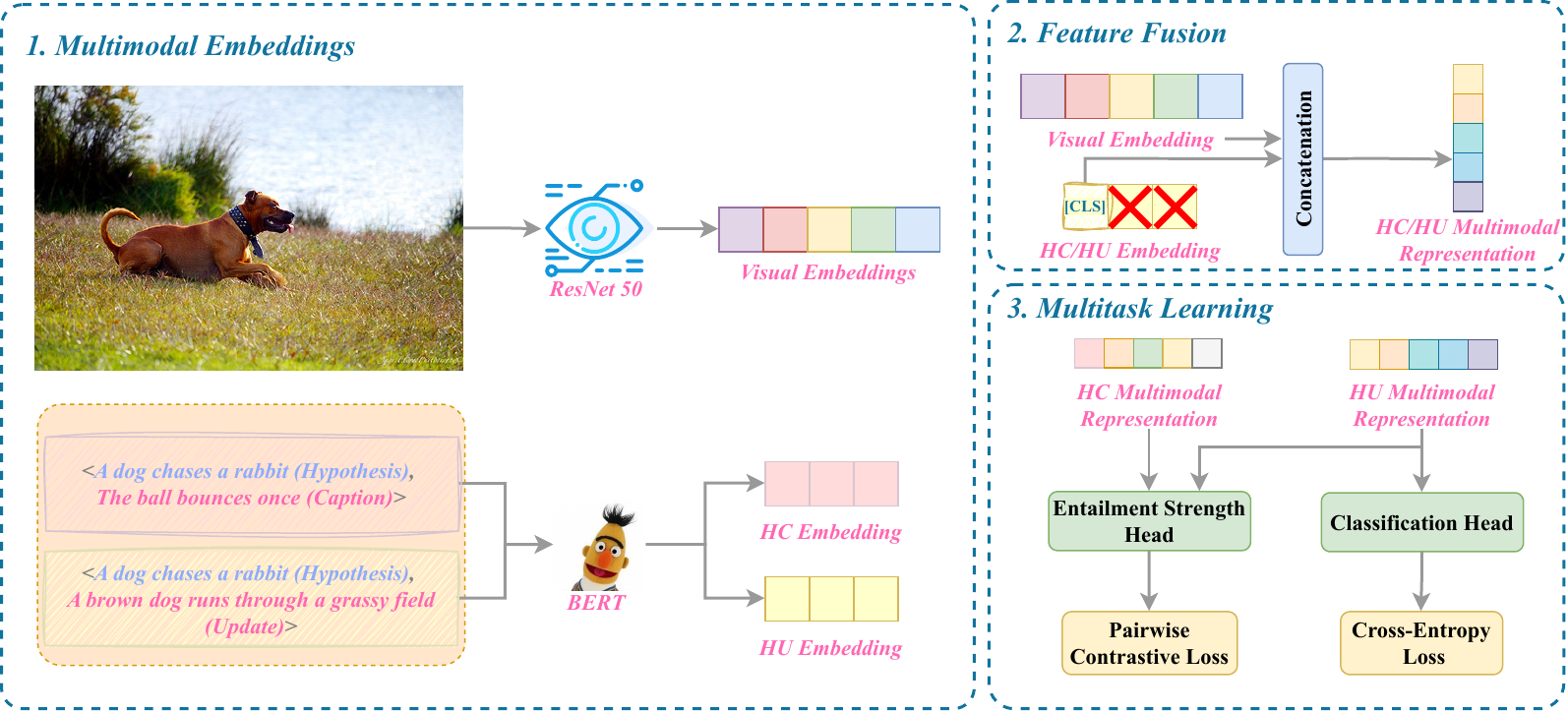}
    \caption{The architecture of our Inference-aware Evaluator, including three modules: Multimodal Embedding, Feature Fusion, and Multitask Learning. HC/HU Embedding means the embedding of the hypothesis-caption/hypothesis-update pair. Similarly, HC/HU Multimodal Representation stands for the multimodel representation of the hypothesis-caption/hypothesis-update pair.}
    \label{fig:model_structure}
\end{figure*}

As mentioned before, for the generation task, we designed a novel reference-free evaluation method that leverages contrastive learning to capture the impact of updates on inference strength. Our model consists of the following components. The overall architecture of the model is illustrated in Figure \ref{fig:model_structure}, which consists of three modules: Multimodal Embedding, Feature Fusion, and Multitask Learning.

% \begin{figure}[ht]
%     \centering
%     \includegraphics[width=\linewidth]{AAAI_EVAL.pdf}
%     \caption{The overall architecture of the model}
%     \label{fig:model_structure}
% \end{figure}

\subsubsection{Multimodal Embedding}
The input data for the model consists of both images and text. To feed the multimodal data into our model, we first get the embeddings of the text and image as follows.

% structured as follows:

\textbf{\textit{Visual Embedding}} Since ResNet \cite{DBLP:conf/cvpr/HeZRS16} has shown great success on vision tasks, such as image classification \cite{DBLP:journals/ijcv/RussakovskyDSKS15, krizhevsky2009learning}, object detection \cite{DBLP:journals/ijcv/EveringhamGWWZ10, DBLP:conf/eccv/LinMBHPRDZ14}, semantic segmentation \cite{DBLP:conf/cvpr/ZhouZPFB017}, we also use it to extract the visual embedding. Specifically, we use the pretrained ResNet-50 model to extract the visual embedding as follows,
% We use Visual features are extracted from each image using a pretrained ResNet50 model. Specifically, the image embedding $f_{\mathcal{I}}$ for an image $\mathcal{I}$ is obtained as follows:
\begin{equation}
\mathbf{i} = \text{ResNet}({I}),
\end{equation}
where $\mathbf{i} \in \mathbb{R}^{d_1}$ denotes the image embedding of the image premise $I$. $d_1$ is the embedding size. $\text{ResNet}(\cdot)$ refers to the ResNet-50 model.
% and $\Theta_{\text{resnet}}$ refers to the parameters of the ResNet50 model.

\textbf{\textit{Texual Embedding}} It is known that BERT \cite{DBLP:conf/naacl/DevlinCLT19} achieves superior performance on various natural language models, such as Language Understanding \cite{DBLP:conf/iclr/WangSMHLB19}, Question Answering \cite{DBLP:conf/emnlp/RajpurkarZLL16} and Commonsense Inference \cite{DBLP:conf/emnlp/ZellersBSC18}. Therefore, we use BERT to extract the textual features. In particular, we encode a pair of text inputs: the hypothesis and update with BERT as follows,
\begin{equation}
\mathbf{e} = \text{BERT}([H, U]),
\end{equation}
where $\mathbf{e} \in \mathbb{R}^{d_2}$ represent the embedding of the [CLS] token output from BERT, which is used to represent the overall semantics of the text pairs. $\text{BERT}(\cdot)$ refers to the BERT model.

% Text features are extracted using a BERT encoder. Specifically, we encode pairs of text inputs: the image caption and hypothesis, and the hypothesis and update. The image caption $\mathcal{C}$ is the textual description of the image $\mathcal{I}$, and the hypothesis $\mathcal{H}$ is the text statement being evaluated. The update $\mathcal{U}$ is the new information that can potentially strengthen or weaken the hypothesis. The BERT encoder processes these pairs to produce embeddings $e_{\mathcal{CH}}$ and $e_{\mathcal{HU}}$, respectively. The [CLS] token output from BERT is used to represent the overall semantics of the text pairs:
% \begin{equation}
% e_{\mathcal{CH}} = \text{BERT}([\mathcal{C}, \mathcal{H}]; \Theta_{\text{bert}}),
% \end{equation}
% \begin{equation}
% e_{\mathcal{HU}} = \text{BERT}([\mathcal{H}, \mathcal{U}]; \Theta_{\text{bert}}),
% \end{equation}
% where $e_{\mathcal{CH}}$ and $e_{\mathcal{HU}} \in \mathbb{R}^d$ represent the embeddings of the text pairs, and $\Theta_{\text{bert}}$ refers to the parameters of the BERT model.

\subsubsection{Feature Fusion} We propose to concatenate the extracted visual and textual features to form a combined multimodal feature representation, denoted as $\mathbf{m}$: 
\begin{equation}
\mathbf{m} = [\mathbf{i}, \mathbf{e}], \label{eq:feature}
\end{equation}
where $[,]$ denotes the concatenation operation. In this context, $\mathbf{m} \in \mathbb{R} ^ {d_1 + d_2}$ represents the combined features that integrate the multimodal information, enabling the model to leverage both visual and textual contexts effectively.

\subsubsection{Multitask Learning}
Our evaluator employs a multitask learning framework to jointly perform classification and inference strength tasks, utilizing shared representations to improve overall performance. The inference strength score is ultimately used to represent the strength of visual entailment brought by updates.

\textbf{\textit{Pairwise Contrastive Learning}}
Since the existing entailment datasets only label update classes without indicating entailment strength, they cannot be used to train a model that predicts this strength for a given (premise, hypothesis, and update) triplet. While human scoring could be an option, it is impractical due to its difficulty, cost, and lack of scalability. Instead, motivated by the contrastive learning framework \cite{DBLP:conf/icml/ChenK0H20}, we develop an unsupervised method to train our evaluator by comparing the entailment strength between pairs, requiring only knowledge of which pair has stronger entailment.
%The existing entailment dataset only labeled the class of updates, but not the entailment strength for the premise, hypothesis, and update. Therefore, we cannot use the existing entailment dataset to train a model that can predict entailment strength for the premise, hypothesis, and update. 
%The simplest method to train a model that can predict entailment strength is to employ people to score for the update, premise, and hypothesis. However, it is hard for people to score a float value that manifests the entitlement strength. In addition, human annotation is costly and is not easy to scale. Motivated by the contrastive learning \cite{DBLP:conf/icml/ChenK0H20}, we devised an unsupervised learning method to train the evaluator. In this setting, we only need to know which one has stronger entailment strength in two pairs (update, premise, and hypothesis) and train the model by entailment strength comparison. 

Specifically, we first devise an entailment strength head to output a numerical score $s$ representing the impact of the update on the hypothesis as follows, 
\begin{equation}
s = \mathbf{W}_s \mathbf{m} + \mathbf{b}_s, \label{eq:score}
\end{equation}
where $\mathbf{W}_s \in \mathbb{R} ^ {d_1 + d_2}$ and $\mathbf{b}_s \in \mathbb{R} ^ 1$ are the trainable weights and bias of the entailment strength layer respectively. 
The entailment strength score $s$ is used as the final measure of the strength of entailment inference, indicating how the update affects the hypothesis. A higher score indicates that the update makes the hypothesis more likely in relation to the premise. In contrast, a lower score indicates that the update makes the hypothesis less likely in relation to the premise.

To train the evaluator, we design a custom pairwise contrastive loss function that can capture the change in entailment strength by comparing triplets (update, premise, and hypothesis).
It is evident that the the entailment strength of the triplet (strengthener, premise, and hypothesis) is bigger than the triplet (caption, premise, and hypothesis) and the the entailment strength of the triplet (weakener, premise, and hypothesis) is smaller than the triplet (caption, premise, and hypothesis). Therefore, we devise the pairwise contrastive loss function as follows,
% To train the evaluator, we design a custom pairwise contrastive loss function to guide the model training by capturing the change in inference strength through the comparison of hypothesis and premise scores with hypothesis and update scores. This allows the model to learn the strength of the update by contrasting these scores. The pairwise contrastive loss function is defined as follows:
\begin{equation}
\mathcal{L}_p = - \frac{1}{N} \sum_{i=1}^N \log \left( \sigma((s_u^i - s_{c}^i) \cdot l^i) \right),
\end{equation}
where $s_{u}^i$ is the score computed by the Eqn.($\ref{eq:score}$) for the triplet (update, premise, and hypothesis). $s_{c}^i$ is the score computed by the Eqn.($\ref{eq:score}$) for the triplet (caption, premise, and hypothesis). 
% and $s{c}^i$ are the scores for the triplet () and the caption for the $i$-th sample, respectively, 
$l^i \in \{-1, 1\}$, where $-1$ represents the update is a weakener and $1$ represents the update is a strengthener. $\sigma(\cdot)$ is the sigmoid function, and $N$ is the number of samples.

\textbf{\textit{Categorical Information Learning}} To further learn the category information of the update, we devise a categorical information loss function. Specifically, we first design a classification head that aims to classify the update as either a strengthener or a weakener as follows, 
\begin{equation}
    \hat{\mathbf{y}} = \sigma(W_c \mathbf{m}_u + b_c),
\end{equation}
where $\sigma$ is the sigmoid activation function.
$W_c \in \mathbb{R} ^ {d_1 + d_2}$ and $b_c \in \mathbb{R} ^ 2$ are the trainable weight and bias of the classification layer. 
$\mathbf{m}_u$ is the combined multimodal feature representation by Eqn.($\ref{eq:feature}$) for the triplet (update, premise, hypothesis). 
$\hat{\mathbf{y}} \in \mathbb{R}^{2}$ is the corresponding predicted label (\ie strengthener and weakener) for the above triplet. 
Thereafter, we utilize the cross-entry loss function to learn the categorical information as follows,
\begin{equation}
    \mathcal{L}_c = -\frac{1}{N} \sum_{i=1}^N \sum_{j=1}^C y_{ij} \log \hat{y}_{ij},
\end{equation}
where $N$ is the number of samples, $C$ is the number of classes, $y_{ij}$ is the ground truth label for the $i$-th sample and the $j$-th class (1 if the sample belongs to the class, otherwise 0), and $\hat{y}_{ij}$ is the predicted probability for the $i$-th sample and the $j$-th class.

% \begin{table}[ht]
%     \centering
%     \begin{tabular}{lccc}
%         \toprule
%         \textbf{Model} & \textbf{Evaluator} & \textbf{BLEU} & \textbf{ROUGEL} \\
%         \midrule
%         \textbf{GPT4o} & & & \\
%         \quad Pearson's Corr & 0.8262 & -0.0081 & -0.1265 \\
%         \quad Spearman's Corr & 0.8037 & -0.0295 & -0.1631 \\
%         \quad Kendall's Tau & 0.6552 & -0.0265 & -0.1180 \\
%         \midrule
%         \textbf{LLaVA-1.5} & & & \\
%         \quad Pearson's Corr & 0.7733 & -0.0777 & -0.1964 \\
%         \quad Spearman's Corr & 0.7368 & -0.0601 & -0.2047 \\
%         \quad Kendall's Tau & 0.6024 & -0.0528 & -0.1502 \\
%         \bottomrule
%     \end{tabular}
% \caption{Correlation coefficients between different metrics and human annotations}
% \label{table:correlation}
% \end{table}

\subsection{Training}
The overall loss function for multitask learning is a weighted sum of the classification loss and the pairwise contrastive loss as follows,
\begin{equation}
\mathcal{L} = (1 - \alpha) \mathcal{L}_p + \alpha \mathcal{L}_c,
\end{equation}
where $\mathcal{L}_c$ is the binary cross-entropy loss for the classification task, $\mathcal{L}_p$ is the pairwise contrastive loss, and $\alpha$ is a hyper-parameter to balance their contributions.
\section{Meta-evaluate Evaluator for Automatic Evaluation}
To verify the effectiveness of our automatic evaluator, we conduct human evaluations on the whole test dataset. 
 For the evaluation model, we select answers from LLaVA-1.5 \cite{DBLP:conf/nips/LiuLWL23a} and GPT-4o.
 More implementation details can be found in the supplementary material.

\subsection{Human Evaluations}
% needs to change in the future.
As we mentioned before, we select LLaVA-1.5 and GPT-4o to generate strengthener and weakener for the 203 images in the testing set, a total $203\times 2 \times 2 = 812$ samples. 
We employed 3 workers for annotation, with each person annotating 812 testing samples.
% using a custom-designed Jupyter Notebook program.
For each test example, we meticulously designed an annotation process to evaluate the scores of the models' generated answers. 
The score was conducted on a 5-point scale, ranging from “weakens a lot” to “strengthens a lot,” with a middle category of “neutral” for updates that have no effect. 
Each worker was paid 15-20 USD per hour. 
After the annotation process, we calculated the inter-annotator agreement rate using Fleiss' $\kappa$, achieving a result of 80.4\%, which involved all annotators. 
This level of concordance among the evaluators suggests that the human evaluation results are reliable.

\subsection{Correlations with Human Evaluations}
We evaluate our proposed metric against traditional metrics commonly used for generation tasks, such as ROUGE-L, BLEU, BERTScore, and CLIPScore. These metrics are widely recognized for their effectiveness in evaluating text generation \cite{DBLP:conf/emnlp/NarayanCL18, DBLP:conf/akbc/LinSZZBCR20} and vision-language tasks \cite{DBLP:conf/eccv/LinMBHPRDZ14, DBLP:conf/eccv/SidorovHRS20}.
% , providing a comprehensive basis for comparison. 
To quantify the alignment between human annotations and model-generated evaluations, we employed three different correlation coefficients: Pearson's $r$, Spearman's $\rho$, and Kendall's $\tau$. 
% Pearson's $r$ measures the linear correlation between two continuous variables, making it suitable for evaluating linear relationships. Spearman's $\rho$ assesses rank correlation, which is ideal for understanding the monotonic relationships between variables, regardless of whether they are linear or not. Kendall's $\tau$ evaluates ordinal association, providing insights into the consistency of pairwise rankings, especially useful for smaller sample sizes and when focusing on the agreement between ranks. These diverse correlation measures offer a comprehensive assessment of how well each metric aligns with human judgments, thus ensuring a robust evaluation of our proposed metric's performance.

\begin{table}[t]
    \centering
    \begin{tabular}{lccc}
        \toprule
        \textbf{Metric} & $r (\%)$ & $\rho (\%)$ & $\tau (\%)$ \\
        \midrule
        \rowcolor{gray!10} \multicolumn{4}{c}{\text{GPT-4o}}  \\
        \quad ROUGE-L & -0.1265 & -0.1631 & -0.1180 \\
        \quad BLEU & -0.0081 & -0.0295 & -0.0265 \\
        \quad BERTScore & -0.0566 & -0.0821 & -0.0558 \\
        \quad CLIPScore & 0.1068 & 0.1179 & 0.0853 \\
        \quad \textbf{Ours} & \textbf{0.8262} & \textbf{0.8037} & \textbf{0.6552} \\
        \midrule
        \rowcolor{gray!10} \multicolumn{4}{c}{\text{LLaVA-1.5}}  \\
        \quad ROUGE-L & -0.1964 & -0.2047 & -0.1502 \\
        \quad BLEU & -0.0777 & -0.0601 & -0.0528 \\
        \quad BERTScore & -0.0950 & -0.1069 & -0.0795 \\
        \quad CLIPScore & 0.2760 & 0.2690 & 0.2035 \\
        \quad \textbf{Ours} & \textbf{0.7733} & \textbf{0.7368} & \textbf{0.6024} \\
        \bottomrule
    \end{tabular}
\caption{Correlation between each evaluation metric and human judgment on VDI, measured by Pearson's $r$, Spearman's $\rho$, and Kendall's $\tau$.  The best metrics for each correlation coefficient are highlighted in bold.}
\label{table:correlation}
\end{table}

\begin{figure*}[t]
    \centering
        % \vspace{-3mm}
\includegraphics[width=\linewidth]{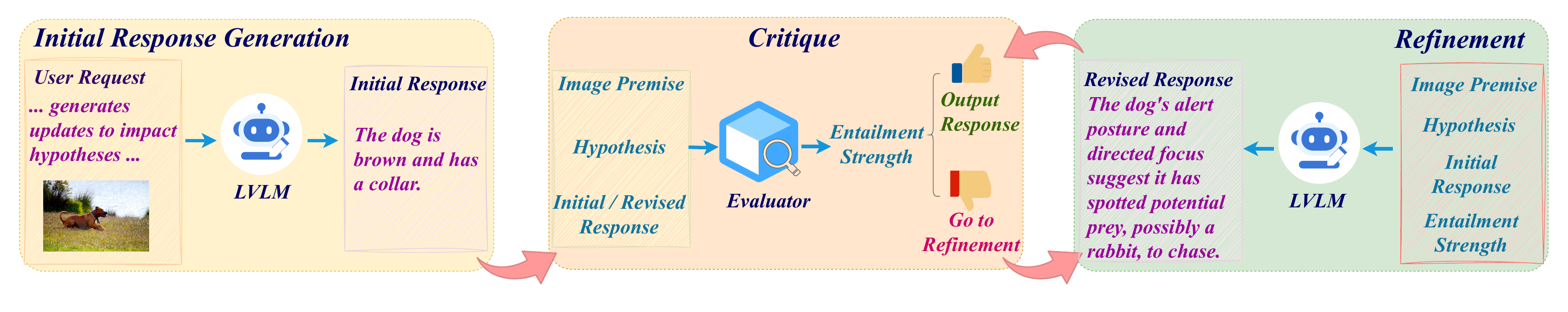}
    \caption{An overview of Reward-driven Update Optimization, which includes three steps: Initial Response Generation, Critique, and Refinement.}
    \vspace{-3mm}
    \label{fig:method}
\end{figure*}

We show the correlation between automatic evaluation and human evaluation in Table \ref{table:correlation}.
% Several observations can be found: 1)
Except for our metric and CLIPScore, other evaluation metrics (\eg ROUGE-L and BLEU) show negative correlations across both GPT-4o and LLaVA-1.5's results. This indicates that these metrics do not align well with human judgments. 
One potential reason is that these metrics pay more attention to the text overlap but this is not suitable for open-domain generation, especially when the answer is not a fixed one.
% modality and cannot process the visual information present in the premise image.
% Although CLIPScore exhibits some degree of positive correlation in the results of GPT-4o and LLaVA-1.5, it is not as strong as our metric. This is due to CLIPScore's inherent design, which primarily focuses on image captioning tasks and cannot perceive the nuanced changes brought by updates in entailment strength.
Notably, our proposed metric consistently outperforms other metrics across all correlation measures, which indicates the effectiveness of our metric. In addition, a key virtue of our method is that compared with some traditional metrics (\eg BLEU and ROUGE-L), our metric is reference-free.
We provide the case study for the evaluator in the supplementary material.

\section{Reward-driven Update Optimization}

In the generation task, our objective is to generate updates based on given premises to either strengthen or weaken hypotheses. In our experiments, we observed that the initial updates may suffer from quality issues, such as simply captioning the images rather than effectively achieving the intended goal. 
% This challenge often arises because the model tends to rely heavily on generating image captions. 
% Especially when there is a significant discrepancy between the image and the hypothesis, the model is more likely to simple captioning rather than generating appropriate updates. 

To address this issue, we propose a new reward-driven update optimization method, which leverages the entailment strength of the generated update. Figure \ref{fig:method} presents an overview of the proposed method. Our method consists of the following steps:

%we incorporated the entailment strength of the generated update into our reward-driven update optimization method. Figure \ref{fig:method} presents an overview of the proposed method. We devise this method based on the following steps:

\begin{enumerate}
    \item \textbf{Initial Response Generation:} We submit the user request to the Large Vision-Language Model (LVLM), to generate the initial response. This response serves as the baseline for subsequent comparisons with the refined responses produced by our method.

    \item \textbf{Critique:} Our inference-aware evaluator serves as the critique, assessing the entailment strength of the generated updates. If the critique assigns a low score, we proceed to the next step (\ie Refinement) to improve the response. We establish a threshold $\eta$ to evaluate the quality of the generated update. Specifically, if the score of a generated strengthener is less than $\eta$ or the score of a generated weakener exceeds $-\eta$, we classify the update as low-quality. Conversely, if the score indicates the update is of high quality, we output the current response as the final result.

    %\item \textbf{Critique:} We use our inference-aware evaluator as the critique, assessing the entailment strength introduced by the generated updates. 
    %After obtaining the score from the critique, if the score indicates low quality, we then resort to the next step (\ie Refinement) to refine the response. Specifically, we set a threshold $\eta$ to judge the quality of the generated update. When the score of the generated strengthener is less than $\eta$ or the score of the generated weakened is bigger than $-\eta$, we deem the generated update as a low-quality generation. 
    %When the score indicates the response is a good update, we then output the current response as the final output.

    % We establish hyper-parameters as thresholds for classifying updates as strengtheners or weakeners.
    \item \textbf{Refinement:} In this step, we feed the score along with the current generation result into the LVLM to refine the response. After generating a new update, we return to the critique step to obtain a new score. This process is repeated until the model produces a high-quality update (as defined in the critique step) or until the loop reaches a maximum iteration count of $M$.

    %\item \textbf{Refinement:} In this step, we  feed the score as well as the current generation result into LVLM to refine the response.  After generating a new update, we go back to the critique step and get a new score. We will continue these steps until the model generates a high-quality update (as we said in the critique step) or the loop achieves a maximum value of $M$.
    % compare the generate result's score with the thresholds set in step 2. For strengtheners, if the score exceeds the threshold, we retain the result; for weakeners, if the score is below the threshold, we retain the result. Otherwise, the model regenerates the updates based on the previous update and score. We also set a hyper-parameter for the number of epochs for regeneration. The loop stops either when the maximum number of epochs is reached or when all updates satisfy the thresholds.
\end{enumerate}

\begin{table}[t]
    \centering
    \resizebox{0.48\textwidth}{!}{
    \begin{tabular}{lcccccc}
        \toprule
        \textbf{Model} & \textbf{ROUGE-L} & \textbf{BLEU}  & \textbf{BERTScore} & \textbf{CLIPScore} & \textbf{Ours} \\
        \midrule
        \rowcolor{gray!10} \multicolumn{6}{c}{\text{Strengthener}}  \\
        \quad InstructBLIP & 0.0601 & 0.0141 & 0.1891 & 0.2111 & 0.7211 \\
        \quad Multimodal-GPT & 0.0541 & 0.0033 & 0.7774 & 0.2426 & 1.0690 \\
        \quad MiniGPT-4 & 0.1376 & 0.0180 & 0.7696 & 0.2705 & 1.4998 \\
        \quad mPLUG-Owl & \textbf{0.3308} & \textbf{0.0781} & 0.8815 & 0.2733 & 2.2887 \\
        \quad LLaVA-1.5 & 0.3163 & 0.0612 & 0.8847 & 0.2788 & 2.7868 \\
        \quad GPT-4o & 0.2702 & 0.0423 & \textbf{0.8954} & 0.2867 & 3.6413 \\
        \quad GPT-4o (Optimized) & 0.2653 & 0.0410 & 0.8787 & \textbf{0.2872} & \textbf{4.0679} \\
        \midrule
        \rowcolor{gray!10} \multicolumn{6}{c}{\text{Weakner}}  \\
        \quad InstructBLIP & 0.0817 & 0.0280 & 0.2614 & 0.2161 & 0.4231 \\
        \quad Multimodal-GPT & 0.0481 & 0.0032 & 0.7748 & 0.2406 & 0.7194 \\
        \quad MiniGPT-4 &  0.1193 & 0.0128 & 0.7183 & 0.2639 & 0.9776 \\
        \quad mPLUG-Owl  & \textbf{0.3386} & \textbf{0.0858} & 0.8842 & 0.2732 & 1.0274 \\
        \quad LLaVA-1.5 & 0.3438 & 0.0773 & 0.8865 & 0.2702 & 0.4834 \\
        \quad GPT-4o & 0.2800 & 0.0451 & \textbf{0.8957} & 0.2782 & -2.5212 \\
        \quad GPT-4o (Optimized) & 0.2768 & 0.0440 & 0.8798 & \textbf{0.2762} & \textbf{-2.9240} \\
        \bottomrule
    \end{tabular}
    }
\caption{Evaluation metrics for strengtheners and weakeners across different models. For our metric, a higher value represents a higher entailment strength brought by updates, while a lower value indicates a lower entailment strength. Therefore, for strengtheners, a higher value reflects a stronger entailment strength update. Conversely, for weakeners, a lower value indicates a more effective weakening update. The best results in each category are highlighted in bold.}
\label{table:metrics}
\end{table}

\section{Evaluate Models on DVE Tasks}
% In this section, we evaluate several vision-language
% models for the Classification Task and Generation Task.

\subsection{Experimental Setup}
% \subsubsection{Classification Task}
% fine-tuning / zero-shot
For the \textit{Classification} Task, we selected seven models, categorized into two types: finetuning-based methods and models evaluated in the zero-shot setting. The finetuning-based models include VILT \cite{DBLP:conf/icml/KimSK21}, FLAVA \cite{DBLP:conf/cvpr/SinghHGCGRK22}, and CLIP \cite{DBLP:conf/icml/RadfordKHRGASAM21}. We fine-tuned these models on our training set with standard cross-entropy classification loss function. 
% VILT uses a unified transformer encoder to jointly process visual and textual inputs.
The models under the zero-shot setting include InstructBLIP \cite{DBLP:conf/nips/Dai0LTZW0FH23}, LLaVA-1.5, mPLUG-Owl \cite{DBLP:journals/corr/abs-2304-14178}, and GPT-4o.
We directly prompt these pretrained LVLMs to generate a prediction for classification results.
% FLAVA combines separate image and text encoders through multimodal pretraining to handle diverse visual-textual interactions. CLIP employs contrastive learning to align visual and textual representations by training on a large dataset of image-text pairs. 
% In the zero-shot setting, models such as InstructBLIP, LLaVA-1.5, mPLUG-Owl, and GPT-4o leverage extensive visual instruction data to enhance their generalization capabilities on unseen data. 
% These models, known for their advanced language understanding and multimodal processing abilities, further diversify our model selection. Most of these models are open-source and widely adopted in the research community, ensuring reproducibility and broad applicability of our findings. 
% \subsubsection{Generation Task}
%hard task, blip can't handle it
For the \textit{Generation} Task, we selected six widely used LVLMs in a zero-shot setting as baselines:
1) InstructBLIP;
2) Multimodal-GPT \cite{DBLP:journals/corr/abs-2305-04790};
3) MiniGPT-4 \cite{DBLP:journals/corr/abs-2304-10592};
4) mPLUG-Owl;
5) LLaVA-1.5;
6) GPT-4o. 
% These LVLMs consist of three fundamental components: a visual encoding module, an alignment mechanism, and a large language model. 
% Each model has been pretrained using curated datasets of visual instruction data. For instance, LLaVA-1.5 utilizes a blend of visual and textual data to enhance its alignment mechanism, thereby achieving superior performance in tasks that necessitate both visual comprehension and textual reasoning. The first five models are open-source and have been extensively adopted in the research community for various vision-language tasks. Although GPT-4o is not open-source, it is included in our evaluation due to its advanced language understanding capabilities and effective multimodal processing abilities. We also test our proposed method in this task.
% \begin{table*}[ht]
% \centering
% \begin{tabular}{|c|c|c|}
% \hline
% \textbf{Model} & \textbf{Strengthener Average Score} & \textbf{Weakener Average Score} \\
% \hline
% InstructBLIP & 0.721 & 0.448 \\
% \hline
% Multimodal-GPT & 1.069 & 0.719 \\
% \hline
% MiniGPT-4 & 1.498 & 0.991 \\
% \hline
% mPLUG-Owl  & 2.311 & 1.040 \\
% \hline
% LLaVA & 2.799 & 0.531 \\
% \hline
% GPT-4 & 3.641 & -2.521 \\
% \hline
% GPT-4 (Agent) & 4.068 & -2.924 \\
% \hline
% \end{tabular}
% \caption{Average scores for Strengthener and Weakener across different models}
% \label{table:scores}
% \end{table*}
We select GPT-4o as the LVLM in reward-driven update optimization. 
More details of the experiments can be found in the supplementary material.

\subsection{Results and Analysis}

\subsubsection{Classification Task}
Table \ref{table:cls} presents the accuracy of the various models on the classification task.
% , where the models predict the category of updates as either strengtheners or weakeners. 
From this table, we observe that: (1) Among the finetuning-based models, CLIP achieves the highest accuracy at 71.10\%, followed by FLAVA at 70.03\%, and VILT at 68.10\%. The likely reason is that the pretraining dataset for CLIP is larger than those used for the other models. %The potential reason is the pertaining data size used for CLIP is bigger than others. 
% The performance of these models is positively correlated with the size of their pretraining datasets. 
% Additionally, CLIP uses contrastive learning to align visual and textual representations effectively. While VILT simplifies model design and improves efficiency, its performance is slightly weaker than the other two models in handling highly complex multimodal relationships. 
(2) The closed-source GPT-4o significantly outperforms all other open-source models with an accuracy of 81.76\%, demonstrating its robust capability. 
% However, other LVLMs do not perform as well on this task. The potential reason could be that GPT-4o has an advanced architecture, extensive multimodal pretraining, and a unified processing approach, enabling it to be the best model for this task. 
(3) The fine-tuning-based models outperform most LVLMs in the zero-shot setting, except for GPT-4o. This suggests that despite being trained on large-scale datasets, current LVLMs still lack sufficient knowledge for our classification task.

%This indicates the existing LVLMs still lack knowledge in our classification task, despite they have been trained in large-scale datasets.
% Except for GPT-4o, finetuning-based models tend to perform better on this classification task
% LLaVA-1.5 performs better than InstructBLIP and mPLUG-Owl because it jointly processes visual and textual inputs, capturing complex interactions more effectively. In contrast, InstructBLIP and mPLUG-Owl handle modalities separately, struggling to fully capture the nuanced relationships between visual and textual data, resulting in relatively lower performance. Overall, these findings suggest that finetuning-based models tend to perform better on this classification task compared to LVLMs in a zero-shot setting. Except for GPT-4o, other models did not demonstrate strong zero-shot reasoning capabilities.

% The performance of these models is positively correlated with the size of their pretraining datasets. Besides, CLIP's superior performance is due to its pretraining on a vast dataset of 400 million image-text pairs and its use of contrastive learning, which effectively aligns visual and textual representations. FLAVA, pretrained on large-scale datasets like ImageNet and COCO, benefits from its multimodal pretraining and fusion strategies. VILT simplifies model design and improves efficiency, but its performance is slightly weaker than CLIP and FLAVA in handling highly complex multimodal relationships.

\begin{table}[t]
    \centering
    \begin{tabular}{lc}
        \toprule
        \textbf{Model} & \textbf{Accuracy (\%)} \\
        \midrule
        % \rowcolor{gray!10} \multicolumn{2}{c}{\text{Finetuning-based Model}}  \\
        \quad VILT & 68.10 \\
        \quad FLAVA & 70.03 \\
        \quad CLIP & 71.10  \\ \midrule
        % \rowcolor{gray!10} \multicolumn{2}{c}{\text{LVLMs}}  \\
        \quad InstructBLIP & 31.32 \\
        \quad mPLUG-Owl  & 31.16 \\
        \quad LLaVA-1.5 & 52.07 \\
        \quad {GPT-4o} & \textbf{81.76} \\
        \bottomrule
    \end{tabular}
    \caption{Performance comparison among different methods in the classification task.}
\label{table:cls}
\end{table}

\subsubsection{Generation Task}
Table \ref{table:metrics} presents the performance of supporter and defeater generation across various assessment metrics. Notably, even GPT-4o does not achieve high scores according to existing generation metrics, highlighting the limitations of current metrics in accurately evaluating the quality of generated updates. This underscores the necessity of our proposed metric.
%Table \ref{table:metrics} shows the performance of the supporter and defeater generation on a wide range of assessment metrics. 
%Notably, even GPT-4o does not achieve a high score based on existing generation metrics. This also suggests that current metrics are insufficient for evaluating the quality of generated updates. This indicates the necessity of our proposed metric.
% From the table, we can see that InstructBLIP performs poorly across all metrics, primarily because it often refuses to generate outputs when tasked with producing weakeners. 
MiniGPT-4 and Multimodal-GPT outperform InstructBLIP in BERTScore, likely due to their more fluent and coherent outputs. This advantage can be attributed to the more advanced language models used by MiniGPT-4 and Multimodal-GPT, which are better equipped to generate contextually appropriate sentences.
%This likely results from both models using more advanced language models that are better at generating coherent and contextually appropriate sentences.
% mPLUG-Owl, despite achieving the best scores in ROUGE-L and BLEU, shows these metrics are slightly negatively correlated with the quality of updates, as indicated in Table \ref{table:correlation}. 
% LLaVA-1.5 demonstrates better generalization than all models except GPT-4o, attributed to its ability to jointly process visual and textual inputs, effectively capturing complex interactions, and its extensive pretraining on large datasets. 
%Among all baselines, GPT-4o achieved the best performance, which demonstrates the robustness of GPT-4o.
%Compared with GPT-4o, our proposed framework GPT-4o (Optimized) shows better. This is because our framework could provide the feedback for GPt-4o and help it refine the low-quality response.

Among all baselines, GPT-4o achieved the best performance, demonstrating its robustness. Our proposed framework, GPT-4o (Optimized), performs even better than GPT-4o alone. This improvement is due to our framework's ability to provide feedback to GPT-4o, enabling it to refine low-quality responses.
Additionally, it is evident that all models perform worse in generating weakeners, with only GPT-4o-based models being able to produce effective weakeners. This is likely because most models tend to default to simple image captioning rather than generating nuanced defeaters.

We also assessed human performance based on our evaluator. The average score for the strengthener is 5.0998, and the weakener score is -4.5412. This demonstrates that there is a significant gap between the model's performance and human performance. 
% Finally, the experimental results show that our reward-driven update optimization model achieves the highest strengthener scores and the lowest weakener scores, indicating its superior performance.
%In addition, we also provide case study for our proposed optimization method in the supplement material.
Finally, we provide a case study of our proposed optimization method in the supplement.%ary material.

% \clearpage
\section{Related Work}
\paragraph{Natural Language Inference}
Textual entailment~\cite{DBLP:conf/emnlp/BowmanAPM15,DBLP:conf/naacl/WilliamsNB18,DBLP:conf/acl/NieWDBWK20}, defined as determining whether a human would typically consider a hypothesis to be likely true given a premise, has become a cornerstone task in natural language processing. 
% This concept gained substantial traction with the release of the SNLI dataset \cite{DBLP:conf/emnlp/BowmanAPM15}, which enabled the training and evaluation of neural models on a large scale. 
% Subsequent datasets, such as the Multi-Genre NLI dataset \cite{DBLP:conf/naacl/WilliamsNB18} and the Adversarial NLI dataset \cite{DBLP:conf/acl/NieWDBWK20}, further expanded the scope and complexity of NLI tasks. 
However, the task of textual entailment has faced criticism, studies have shown significant variability in human agreement on entailment judgments \cite{DBLP:journals/tacl/PavlickK19}, leading to the proposal of alternative approaches that use ordinal or numeric values to represent plausibility \cite{DBLP:journals/tacl/ZhangRDD17, DBLP:conf/acl/DurmeS18, DBLP:conf/acl/ChenJPSD20, DBLP:conf/nodalida/TalmanCVHT23}. This shift aims to capture the nuanced nature of entailment more accurately.
In recent years, the focus has shifted towards the defeasibility of textual entailments, which involves revising or overturning conclusions based on new evidence. The $\delta$-NLI dataset extends existing NLI datasets by including scenarios where new information can alter inferences, providing a more realistic evaluation of models' reasoning abilities \cite{DBLP:conf/emnlp/RudingerSHBFBSC20}. Similarly, the BoardgameQA dataset measures the reasoning capacity of language models when faced with contradictory information, guided by source preferences and implicit background knowledge, better reflecting real-world reasoning challenges \cite{DBLP:conf/nips/KazemiYBKXIR23}.
% These efforts aim to enhance the robustness and applicability of NLI models in dynamic information environments.
However, the defeasible entailment inference in the multimodal setting is still unexplored. 

\paragraph{Visual Understanding Tasks}
Visual Question Answering (VQA), image captioning, and visual reasoning are common visual understanding tasks. VQA aims to answer natural language questions based on provided visual information. The VQA-v1.0 dataset \cite{DBLP:conf/iccv/AntolALMBZP15} was one of the first to address this task, focusing on the basic interaction between visual content and natural language questions. However, it faced issues related to biases and limited reasoning capabilities \cite{DBLP:journals/corr/abs-1901-06706}. To address these limitations, several datasets \cite{DBLP:conf/cvpr/JohnsonHMFZG17, DBLP:conf/cvpr/GoyalKSBP17, han2023coremm, DBLP:conf/wacv/MathewBTKVJ22,DBLP:conf/nips/LuQCXZZYLZ21} have been developed to reduce biases and enhance reasoning capabilities. While VQA focuses on understanding and answering questions about visual content, 
image captioning involves generating natural language descriptions of an image's content \cite{DBLP:conf/eccv/LinMBHPRDZ14,DBLP:journals/tacl/YoungLHH14,DBLP:conf/eccv/SidorovHRS20,cite1,cite2,cite3}. 
% Image captioning benchmarks, such as MS COCO 
% \cite{DBLP:conf/eccv/LinMBHPRDZ14}, Flickr30k \cite{DBLP:journals/tacl/YoungLHH14}, and TextCaps \cite{DBLP:conf/eccv/SidorovHRS20}.
% , provide extensive datasets that challenge models to describe images with detailed and accurate captions. 
In addition, 
% to VQA and image captioning,
visual reasoning involves understanding relationships and interactions between visual elements, enhancing comprehension of visual content~\cite{DBLP:conf/cvpr/ThrushJBSWKR22,DBLP:journals/corr/abs-2310-10207,cite5}. 
% The Winoground dataset \cite{DBLP:conf/cvpr/ThrushJBSWKR22} probes vision and language models for visio-linguistic compositionality, testing their ability to understand and generate compositional phrases. 
% Bongard-OpenWorld \cite{DBLP:journals/corr/abs-2310-10207}, is designed for few-shot reasoning about free-form visual concepts in real-world settings. These datasets collectively advance the capabilities of models in understanding and reasoning about complex visual information.
However, these tasks can not capture fine-grained semantics reasoning relation change brought by the new information.

\section{Conclusion}
In this paper, we present a novel defeasible visual entailment task and a new benchmark for studying defeasibility in visual entailment.
We also propose a novel inference-ware evaluator for capturing the change of entailment strength brought by the update and a new reward-driven update optimization
method to further improve the quality of the update generated by the multimodal model. Our experimental results clearly show the effectiveness of our proposed inference-aware evaluator and reward-driven update optimization
method.

% \section{Limitation}

% \clearpage

\section{Acknowledgments}
We want to thank our anonymous reviewers for their feedback. This work was supported in part by the DARPA Perceptually-Enabled Task Guidance (PTG) Program under contract number HR00112220005, the DARPA Assured Neuro Symbolic Learning and Reasoning (ANSR) Program under contract number HR001122S0039, the National Science Foundation grant IIS-1652835, the AFOSR award FA9550-23-1-0239, and OpenAI Researcher Access Program 0000006384.

\bibliography{aaai25}

\begin{thebibliography}{59}
\providecommand{\natexlab}[1]{#1}

\bibitem[{Antol et~al.(2015)Antol, Agrawal, Lu, Mitchell, Batra, Zitnick, and Parikh}]{DBLP:conf/iccv/AntolALMBZP15}
Antol, S.; Agrawal, A.; Lu, J.; Mitchell, M.; Batra, D.; Zitnick, C.~L.; and Parikh, D. 2015.
\newblock {VQA:} Visual Question Answering.
\newblock In \emph{{ICCV}}, 2425--2433. {IEEE} Computer Society.

\bibitem[{Bos and Markert(2005)}]{DBLP:conf/naacl/BosM05}
Bos, J.; and Markert, K. 2005.
\newblock Recognising Textual Entailment with Logical Inference.
\newblock In \emph{{HLT/EMNLP}}, 628--635. The Association for Computational Linguistics.

\bibitem[{Bowman et~al.(2015)Bowman, Angeli, Potts, and Manning}]{DBLP:conf/emnlp/BowmanAPM15}
Bowman, S.~R.; Angeli, G.; Potts, C.; and Manning, C.~D. 2015.
\newblock A large annotated corpus for learning natural language inference.
\newblock In \emph{{EMNLP}}, 632--642. The Association for Computational Linguistics.

\bibitem[{Chang et~al.(2024)Chang, Jing, Zhang, and Zhang}]{cite2}
Chang, Y.; Jing, L.; Zhang, X.; and Zhang, Y. 2024.
\newblock A Unified Hallucination Mitigation Framework for Large Vision-Language Models.
\newblock \emph{CoRR}, abs/2409.16494.

\bibitem[{Chen et~al.(2020{\natexlab{a}})Chen, Jiang, Poliak, Sakaguchi, and Durme}]{DBLP:conf/acl/ChenJPSD20}
Chen, T.; Jiang, Z.; Poliak, A.; Sakaguchi, K.; and Durme, B.~V. 2020{\natexlab{a}}.
\newblock Uncertain Natural Language Inference.
\newblock In \emph{{ACL}}, 8772--8779. Association for Computational Linguistics.

\bibitem[{Chen et~al.(2020{\natexlab{b}})Chen, Kornblith, Norouzi, and Hinton}]{DBLP:conf/icml/ChenK0H20}
Chen, T.; Kornblith, S.; Norouzi, M.; and Hinton, G.~E. 2020{\natexlab{b}}.
\newblock A Simple Framework for Contrastive Learning of Visual Representations.
\newblock In \emph{{ICML}}, 1597--1607. {PMLR}.

\bibitem[{Chen et~al.(2020{\natexlab{c}})Chen, Li, Yu, Kholy, Ahmed, Gan, Cheng, and Liu}]{DBLP:conf/eccv/ChenLYK0G0020}
Chen, Y.; Li, L.; Yu, L.; Kholy, A.~E.; Ahmed, F.; Gan, Z.; Cheng, Y.; and Liu, J. 2020{\natexlab{c}}.
\newblock {UNITER:} UNiversal Image-TExt Representation Learning.
\newblock In \emph{{ECCV}}, 104--120. Springer.

\bibitem[{Cui et~al.(2024)Cui, Milikic, Feng, Ismayilzada, Paul, Bosselut, and Faltings}]{DBLP:journals/corr/abs-2401-03183}
Cui, S.; Milikic, L.; Feng, Y.; Ismayilzada, M.; Paul, D.; Bosselut, A.; and Faltings, B. 2024.
\newblock {\(\delta\)}-CAUSAL: Exploring Defeasibility in Causal Reasoning.
\newblock \emph{CoRR}, abs/2401.03183.

\bibitem[{Dagan, Glickman, and Magnini(2005)}]{DBLP:conf/mlcw/DaganGM05}
Dagan, I.; Glickman, O.; and Magnini, B. 2005.
\newblock The {PASCAL} Recognising Textual Entailment Challenge.
\newblock In \emph{{MLCW}}, volume 3944, 177--190. Springer.

\bibitem[{Dai et~al.(2023)Dai, Li, Li, Tiong, Zhao, Wang, Li, Fung, and Hoi}]{DBLP:conf/nips/Dai0LTZW0FH23}
Dai, W.; Li, J.; Li, D.; Tiong, A. M.~H.; Zhao, J.; Wang, W.; Li, B.; Fung, P.; and Hoi, S. C.~H. 2023.
\newblock InstructBLIP: Towards General-purpose Vision-Language Models with Instruction Tuning.
\newblock In \emph{NeurIPS}.

\bibitem[{Devlin et~al.(2019)Devlin, Chang, Lee, and Toutanova}]{DBLP:conf/naacl/DevlinCLT19}
Devlin, J.; Chang, M.; Lee, K.; and Toutanova, K. 2019.
\newblock {BERT:} Pre-training of Deep Bidirectional Transformers for Language Understanding.
\newblock In \emph{{NAACL-HLT}}, 4171--4186. Association for Computational Linguistics.

\bibitem[{Everingham et~al.(2010)Everingham, Gool, Williams, Winn, and Zisserman}]{DBLP:journals/ijcv/EveringhamGWWZ10}
Everingham, M.; Gool, L.~V.; Williams, C. K.~I.; Winn, J.~M.; and Zisserman, A. 2010.
\newblock The Pascal Visual Object Classes {(VOC)} Challenge.
\newblock \emph{IJCV}, 88: 303--338.

\bibitem[{Gong et~al.(2023)Gong, Lyu, Zhang, Wang, Zheng, Zhao, Liu, Zhang, Luo, and Chen}]{DBLP:journals/corr/abs-2305-04790}
Gong, T.; Lyu, C.; Zhang, S.; Wang, Y.; Zheng, M.; Zhao, Q.; Liu, K.; Zhang, W.; Luo, P.; and Chen, K. 2023.
\newblock MultiModal-GPT: {A} Vision and Language Model for Dialogue with Humans.
\newblock \emph{CoRR}, abs/2305.04790.

\bibitem[{Goyal et~al.(2017)Goyal, Khot, Summers{-}Stay, Batra, and Parikh}]{DBLP:conf/cvpr/GoyalKSBP17}
Goyal, Y.; Khot, T.; Summers{-}Stay, D.; Batra, D.; and Parikh, D. 2017.
\newblock Making the {V} in {VQA} Matter: Elevating the Role of Image Understanding in Visual Question Answering.
\newblock In \emph{{CVPR}}, 6325--6334. {IEEE} Computer Society.

\bibitem[{Han et~al.(2023)Han, You, Liu, Chen, Zheng, Mrini, Lin, Wang, Zhai, Yuan, Wang, and Yang}]{han2023coremm}
Han, X.; You, Q.; Liu, Y.; Chen, W.; Zheng, H.; Mrini, K.; Lin, X.; Wang, Y.; Zhai, B.; Yuan, J.; Wang, H.; and Yang, H. 2023.
\newblock InfiMM-Eval: Complex Open-Ended Reasoning Evaluation For Multi-Modal Large Language Models.

\bibitem[{He et~al.(2016)He, Zhang, Ren, and Sun}]{DBLP:conf/cvpr/HeZRS16}
He, K.; Zhang, X.; Ren, S.; and Sun, J. 2016.
\newblock Deep Residual Learning for Image Recognition.
\newblock In \emph{{CVPR}}. {IEEE} Computer Society.

\bibitem[{Hessel et~al.(2021)Hessel, Holtzman, Forbes, Bras, and Choi}]{DBLP:conf/emnlp/HesselHFBC21}
Hessel, J.; Holtzman, A.; Forbes, M.; Bras, R.~L.; and Choi, Y. 2021.
\newblock CLIPScore: {A} Reference-free Evaluation Metric for Image Captioning.
\newblock In \emph{{EMNLP}}. Association for Computational Linguistics.

\bibitem[{Jing and Du(2024)}]{cite4}
Jing, L.; and Du, X. 2024.
\newblock {FGAIF:} Aligning Large Vision-Language Models with Fine-grained {AI} Feedback.
\newblock \emph{CoRR}, abs/2404.05046.

\bibitem[{Jing et~al.(2024)Jing, Li, Chen, and Du}]{cite1}
Jing, L.; Li, R.; Chen, Y.; and Du, X. 2024.
\newblock FaithScore: Fine-grained Evaluations of Hallucinations in Large Vision-Language Models.
\newblock In Al{-}Onaizan, Y.; Bansal, M.; and Chen, Y., eds., \emph{Findings of the Association for Computational Linguistics: {EMNLP} 2024, Miami, Florida, USA, November 12-16, 2024}, 5042--5063. Association for Computational Linguistics.

\bibitem[{Jing, Zuo, and Zhang(2024)}]{cite3}
Jing, L.; Zuo, J.; and Zhang, Y. 2024.
\newblock Fine-grained and Explainable Factuality Evaluation for Multimodal Summarization.
\newblock \emph{CoRR}, abs/2402.11414.

\bibitem[{Johnson et~al.(2017)Johnson, Hariharan, van~der Maaten, Fei{-}Fei, Zitnick, and Girshick}]{DBLP:conf/cvpr/JohnsonHMFZG17}
Johnson, J.; Hariharan, B.; van~der Maaten, L.; Fei{-}Fei, L.; Zitnick, C.~L.; and Girshick, R.~B. 2017.
\newblock {CLEVR:} {A} Diagnostic Dataset for Compositional Language and Elementary Visual Reasoning.
\newblock In \emph{{CVPR}}, 1988--1997. {IEEE} Computer Society.

\bibitem[{Kazemi et~al.(2023)Kazemi, Yuan, Bhatia, Kim, Xu, Imbrasaite, and Ramachandran}]{DBLP:conf/nips/KazemiYBKXIR23}
Kazemi, M.; Yuan, Q.; Bhatia, D.; Kim, N.; Xu, X.; Imbrasaite, V.; and Ramachandran, D. 2023.
\newblock BoardgameQA: {A} Dataset for Natural Language Reasoning with Contradictory Information.
\newblock In \emph{NeurIPS 2023}.

\bibitem[{Kim, Son, and Kim(2021)}]{DBLP:conf/icml/KimSK21}
Kim, W.; Son, B.; and Kim, I. 2021.
\newblock ViLT: Vision-and-Language Transformer Without Convolution or Region Supervision.
\newblock In \emph{{ICML}}, volume 139 of \emph{Proceedings of Machine Learning Research}, 5583--5594. {PMLR}.

\bibitem[{Kingma and Ba(2015)}]{DBLP:journals/corr/KingmaB14}
Kingma, D.~P.; and Ba, J. 2015.
\newblock Adam: {A} Method for Stochastic Optimization.
\newblock In \emph{{ICLR}}.

\bibitem[{Krizhevsky, Hinton et~al.(2009)}]{krizhevsky2009learning}
Krizhevsky, A.; Hinton, G.; et~al. 2009.
\newblock Learning multiple layers of features from tiny images.
\newblock Technical report, Toronto, ON, Canada.

\bibitem[{Lin et~al.(2020)Lin, Shen, Zhou, Zhou, Bhagavatula, Choi, and Ren}]{DBLP:conf/akbc/LinSZZBCR20}
Lin, B.~Y.; Shen, M.; Zhou, W.; Zhou, P.; Bhagavatula, C.; Choi, Y.; and Ren, X. 2020.
\newblock CommonGen: {A} Constrained Text Generation Challenge for Generative Commonsense Reasoning.
\newblock In \emph{{AKBC}}.

\bibitem[{Lin(2004)}]{lin2004rouge}
Lin, C.-Y. 2004.
\newblock Rouge: A package for automatic evaluation of summaries.
\newblock In \emph{Text summarization branches out}, 74--81.

\bibitem[{Lin et~al.(2014)Lin, Maire, Belongie, Hays, Perona, Ramanan, Doll{\'{a}}r, and Zitnick}]{DBLP:conf/eccv/LinMBHPRDZ14}
Lin, T.; Maire, M.; Belongie, S.~J.; Hays, J.; Perona, P.; Ramanan, D.; Doll{\'{a}}r, P.; and Zitnick, C.~L. 2014.
\newblock Microsoft {COCO:} Common Objects in Context.
\newblock In \emph{{ECCV} 2014}, 740--755. Springer.

\bibitem[{Liu et~al.(2023)Liu, Li, Wu, and Lee}]{DBLP:conf/nips/LiuLWL23a}
Liu, H.; Li, C.; Wu, Q.; and Lee, Y.~J. 2023.
\newblock Visual Instruction Tuning.
\newblock In \emph{NeurIPS}.

\bibitem[{Lu et~al.(2021)Lu, Qiu, Chen, Xia, Zhao, Zhang, Yu, Liang, and Zhu}]{DBLP:conf/nips/LuQCXZZYLZ21}
Lu, P.; Qiu, L.; Chen, J.; Xia, T.; Zhao, Y.; Zhang, W.; Yu, Z.; Liang, X.; and Zhu, S. 2021.
\newblock IconQA: {A} New Benchmark for Abstract Diagram Understanding and Visual Language Reasoning.
\newblock In \emph{NeurIPS}.

\bibitem[{MacCartney and Manning(2009)}]{DBLP:conf/iwcs/MacCartneyM09}
MacCartney, B.; and Manning, C.~D. 2009.
\newblock An extended model of natural logic.
\newblock In \emph{{IWCS}}, 140--156. Association for Computational Linguistics.

\bibitem[{Mathew et~al.(2022)Mathew, Bagal, Tito, Karatzas, Valveny, and Jawahar}]{DBLP:conf/wacv/MathewBTKVJ22}
Mathew, M.; Bagal, V.; Tito, R.; Karatzas, D.; Valveny, E.; and Jawahar, C.~V. 2022.
\newblock InfographicVQA.
\newblock In \emph{{WACV}}, 2582--2591. {IEEE}.

\bibitem[{Narayan, Cohen, and Lapata(2018)}]{DBLP:conf/emnlp/NarayanCL18}
Narayan, S.; Cohen, S.~B.; and Lapata, M. 2018.
\newblock Don't Give Me the Details, Just the Summary! Topic-Aware Convolutional Neural Networks for Extreme Summarization.
\newblock In \emph{EMNLP}, 1797--1807. Association for Computational Linguistics.

\bibitem[{Nie et~al.(2020)Nie, Williams, Dinan, Bansal, Weston, and Kiela}]{DBLP:conf/acl/NieWDBWK20}
Nie, Y.; Williams, A.; Dinan, E.; Bansal, M.; Weston, J.; and Kiela, D. 2020.
\newblock Adversarial {NLI:} {A} New Benchmark for Natural Language Understanding.
\newblock In \emph{{ACL}}, 4885--4901. Association for Computational Linguistics.

\bibitem[{Papineni et~al.(2002)Papineni, Roukos, Ward, and Zhu}]{DBLP:conf/acl/PapineniRWZ02}
Papineni, K.; Roukos, S.; Ward, T.; and Zhu, W. 2002.
\newblock Bleu: a Method for Automatic Evaluation of Machine Translation.
\newblock In \emph{ACL}, 311--318. {ACL}.

\bibitem[{Pavlick and Kwiatkowski(2019)}]{DBLP:journals/tacl/PavlickK19}
Pavlick, E.; and Kwiatkowski, T. 2019.
\newblock Inherent Disagreements in Human Textual Inferences.
\newblock \emph{TACL}, 7: 677--694.

\bibitem[{Qiao et~al.(2023)Qiao, Jing, Song, Chen, Zhu, and Nie}]{cite5}
Qiao, Y.; Jing, L.; Song, X.; Chen, X.; Zhu, L.; and Nie, L. 2023.
\newblock Mutual-Enhanced Incongruity Learning Network for Multi-Modal Sarcasm Detection.
\newblock In Williams, B.; Chen, Y.; and Neville, J., eds., \emph{Thirty-Seventh {AAAI} Conference on Artificial Intelligence, {AAAI} 2023, Thirty-Fifth Conference on Innovative Applications of Artificial Intelligence, {IAAI} 2023, Thirteenth Symposium on Educational Advances in Artificial Intelligence, {EAAI} 2023, Washington, DC, USA, February 7-14, 2023}, 9507--9515. {AAAI} Press.

\bibitem[{Radford et~al.(2021)Radford, Kim, Hallacy, Ramesh, Goh, Agarwal, Sastry, Askell, Mishkin, Clark, Krueger, and Sutskever}]{DBLP:conf/icml/RadfordKHRGASAM21}
Radford, A.; Kim, J.~W.; Hallacy, C.; Ramesh, A.; Goh, G.; Agarwal, S.; Sastry, G.; Askell, A.; Mishkin, P.; Clark, J.; Krueger, G.; and Sutskever, I. 2021.
\newblock Learning Transferable Visual Models From Natural Language Supervision.
\newblock In \emph{{ICML}}, volume 139, 8748--8763. {PMLR}.

\bibitem[{Rajpurkar et~al.(2016)Rajpurkar, Zhang, Lopyrev, and Liang}]{DBLP:conf/emnlp/RajpurkarZLL16}
Rajpurkar, P.; Zhang, J.; Lopyrev, K.; and Liang, P. 2016.
\newblock SQuAD: 100, 000+ Questions for Machine Comprehension of Text.
\newblock In \emph{{EMNLP}}, 2383--2392. The Association for Computational Linguistics.

\bibitem[{Rudinger et~al.(2020)Rudinger, Shwartz, Hwang, Bhagavatula, Forbes, Bras, Smith, and Choi}]{DBLP:conf/emnlp/RudingerSHBFBSC20}
Rudinger, R.; Shwartz, V.; Hwang, J.~D.; Bhagavatula, C.; Forbes, M.; Bras, R.~L.; Smith, N.~A.; and Choi, Y. 2020.
\newblock Thinking Like a Skeptic: Defeasible Inference in Natural Language.
\newblock In \emph{{EMNLP}}, 4661--4675. Association for Computational Linguistics.

\bibitem[{Russakovsky et~al.(2015)Russakovsky, Deng, Su, Krause, Satheesh, Ma, Huang, Karpathy, Khosla, Bernstein, Berg, and Fei{-}Fei}]{DBLP:journals/ijcv/RussakovskyDSKS15}
Russakovsky, O.; Deng, J.; Su, H.; Krause, J.; Satheesh, S.; Ma, S.; Huang, Z.; Karpathy, A.; Khosla, A.; Bernstein, M.~S.; Berg, A.~C.; and Fei{-}Fei, L. 2015.
\newblock ImageNet Large Scale Visual Recognition Challenge.
\newblock \emph{IJCV}, 115: 211--252.

\bibitem[{Sakaguchi and Durme(2018)}]{DBLP:conf/acl/DurmeS18}
Sakaguchi, K.; and Durme, B.~V. 2018.
\newblock Efficient Online Scalar Annotation with Bounded Support.
\newblock In \emph{{ACL}}, 208--218. Association for Computational Linguistics.

\bibitem[{Sidorov et~al.(2020)Sidorov, Hu, Rohrbach, and Singh}]{DBLP:conf/eccv/SidorovHRS20}
Sidorov, O.; Hu, R.; Rohrbach, M.; and Singh, A. 2020.
\newblock TextCaps: {A} Dataset for Image Captioning with Reading Comprehension.
\newblock In \emph{{ECCV}}, 742--758. Springer.

\bibitem[{Singh et~al.(2022)Singh, Hu, Goswami, Couairon, Galuba, Rohrbach, and Kiela}]{DBLP:conf/cvpr/SinghHGCGRK22}
Singh, A.; Hu, R.; Goswami, V.; Couairon, G.; Galuba, W.; Rohrbach, M.; and Kiela, D. 2022.
\newblock {FLAVA:} {A} Foundational Language And Vision Alignment Model.
\newblock In \emph{{CVPR}}, 15617--15629. {IEEE}.

\bibitem[{Talman et~al.(2023)Talman, {\c{C}}elikkanat, Virpioja, Heinonen, and Tiedemann}]{DBLP:conf/nodalida/TalmanCVHT23}
Talman, A.; {\c{C}}elikkanat, H.; Virpioja, S.; Heinonen, M.; and Tiedemann, J. 2023.
\newblock Uncertainty-Aware Natural Language Inference with Stochastic Weight Averaging.
\newblock In Alum{\"{a}}e, T.; and Fishel, M., eds., \emph{NoDaLiDa}, 358--365. University of Tartu Library.

\bibitem[{Thrush et~al.(2022)Thrush, Jiang, Bartolo, Singh, Williams, Kiela, and Ross}]{DBLP:conf/cvpr/ThrushJBSWKR22}
Thrush, T.; Jiang, R.; Bartolo, M.; Singh, A.; Williams, A.; Kiela, D.; and Ross, C. 2022.
\newblock Winoground: Probing Vision and Language Models for Visio-Linguistic Compositionality.
\newblock In \emph{{CVPR}}, 5228--5238. {IEEE}.

\bibitem[{Wang et~al.(2019)Wang, Singh, Michael, Hill, Levy, and Bowman}]{DBLP:conf/iclr/WangSMHLB19}
Wang, A.; Singh, A.; Michael, J.; Hill, F.; Levy, O.; and Bowman, S.~R. 2019.
\newblock {GLUE:} {A} Multi-Task Benchmark and Analysis Platform for Natural Language Understanding.
\newblock In \emph{{ICLR}}. OpenReview.net.

\bibitem[{Wang et~al.(2022)Wang, Yang, Men, Lin, Bai, Li, Ma, Zhou, Zhou, and Yang}]{DBLP:conf/icml/WangYMLBLMZZY22}
Wang, P.; Yang, A.; Men, R.; Lin, J.; Bai, S.; Li, Z.; Ma, J.; Zhou, C.; Zhou, J.; and Yang, H. 2022.
\newblock {OFA:} Unifying Architectures, Tasks, and Modalities Through a Simple Sequence-to-Sequence Learning Framework.
\newblock In \emph{{ICML}}, 23318--23340. {PMLR}.

\bibitem[{Williams, Nangia, and Bowman(2018)}]{DBLP:conf/naacl/WilliamsNB18}
Williams, A.; Nangia, N.; and Bowman, S.~R. 2018.
\newblock A Broad-Coverage Challenge Corpus for Sentence Understanding through Inference.
\newblock In \emph{{NAACL-HLT}}, 1112--1122. Association for Computational Linguistics.

\bibitem[{Wu et~al.(2023)Wu, Ma, Li, Wang, Zhang, Zhu, and Wang}]{DBLP:journals/corr/abs-2310-10207}
Wu, R.; Ma, X.; Li, Q.; Wang, W.; Zhang, Z.; Zhu, S.; and Wang, Y. 2023.
\newblock Bongard-OpenWorld: Few-Shot Reasoning for Free-form Visual Concepts in the Real World.
\newblock \emph{CoRR}, abs/2310.10207.

\bibitem[{Xie et~al.(2019)Xie, Lai, Doran, and Kadav}]{DBLP:journals/corr/abs-1901-06706}
Xie, N.; Lai, F.; Doran, D.; and Kadav, A. 2019.
\newblock Visual Entailment: {A} Novel Task for Fine-Grained Image Understanding.
\newblock \emph{CoRR}, abs/1901.06706.

\bibitem[{Ye et~al.(2023)Ye, Xu, Xu, Ye, Yan, Zhou, Wang, Hu, Shi, Shi, Li, Xu, Chen, Tian, Qi, Zhang, and Huang}]{DBLP:journals/corr/abs-2304-14178}
Ye, Q.; Xu, H.; Xu, G.; Ye, J.; Yan, M.; Zhou, Y.; Wang, J.; Hu, A.; Shi, P.; Shi, Y.; Li, C.; Xu, Y.; Chen, H.; Tian, J.; Qi, Q.; Zhang, J.; and Huang, F. 2023.
\newblock mPLUG-Owl: Modularization Empowers Large Language Models with Multimodality.
\newblock \emph{CoRR}, abs/2304.14178.

\bibitem[{Young et~al.(2014)Young, Lai, Hodosh, and Hockenmaier}]{DBLP:journals/tacl/YoungLHH14}
Young, P.; Lai, A.; Hodosh, M.; and Hockenmaier, J. 2014.
\newblock From image descriptions to visual denotations: New similarity metrics for semantic inference over event descriptions.
\newblock \emph{TACL}, 2: 67--78.

\bibitem[{Yu et~al.(2022)Yu, Wang, Vasudevan, Yeung, Seyedhosseini, and Wu}]{DBLP:journals/tmlr/YuWVYSW22}
Yu, J.; Wang, Z.; Vasudevan, V.; Yeung, L.; Seyedhosseini, M.; and Wu, Y. 2022.
\newblock CoCa: Contrastive Captioners are Image-Text Foundation Models.
\newblock \emph{TMLR}.

\bibitem[{Zellers et~al.(2018)Zellers, Bisk, Schwartz, and Choi}]{DBLP:conf/emnlp/ZellersBSC18}
Zellers, R.; Bisk, Y.; Schwartz, R.; and Choi, Y. 2018.
\newblock {SWAG:} {A} Large-Scale Adversarial Dataset for Grounded Commonsense Inference.
\newblock In \emph{EMNLP}, 93--104. Association for Computational Linguistics.

\bibitem[{Zhang et~al.(2017)Zhang, Rudinger, Duh, and Durme}]{DBLP:journals/tacl/ZhangRDD17}
Zhang, S.; Rudinger, R.; Duh, K.; and Durme, B.~V. 2017.
\newblock Ordinal Common-sense Inference.
\newblock \emph{TACL}, 5: 379--395.

\bibitem[{Zhang et~al.(2020)Zhang, Kishore, Wu, Weinberger, and Artzi}]{DBLP:conf/iclr/ZhangKWWA20}
Zhang, T.; Kishore, V.; Wu, F.; Weinberger, K.~Q.; and Artzi, Y. 2020.
\newblock BERTScore: Evaluating Text Generation with {BERT}.
\newblock In \emph{{ICLR}}. OpenReview.net.

\bibitem[{Zhou et~al.(2017)Zhou, Zhao, Puig, Fidler, Barriuso, and Torralba}]{DBLP:conf/cvpr/ZhouZPFB017}
Zhou, B.; Zhao, H.; Puig, X.; Fidler, S.; Barriuso, A.; and Torralba, A. 2017.
\newblock Scene Parsing through {ADE20K} Dataset.
\newblock In \emph{{CVPR}}, 5122--5130. {IEEE} Computer Society.

\bibitem[{Zhu et~al.(2023)Zhu, Chen, Shen, Li, and Elhoseiny}]{DBLP:journals/corr/abs-2304-10592}
Zhu, D.; Chen, J.; Shen, X.; Li, X.; and Elhoseiny, M. 2023.
\newblock MiniGPT-4: Enhancing Vision-Language Understanding with Advanced Large Language Models.
\newblock \emph{CoRR}, abs/2304.10592.

\end{thebibliography}
\clearpage
\appendix

\section{Prompts}
In this section, we present all the prompts we used in this paper.
\subsection{Prompt for the Classification Task}
\begin{figure}[ht]
    \centering
    \includegraphics[width=\linewidth]{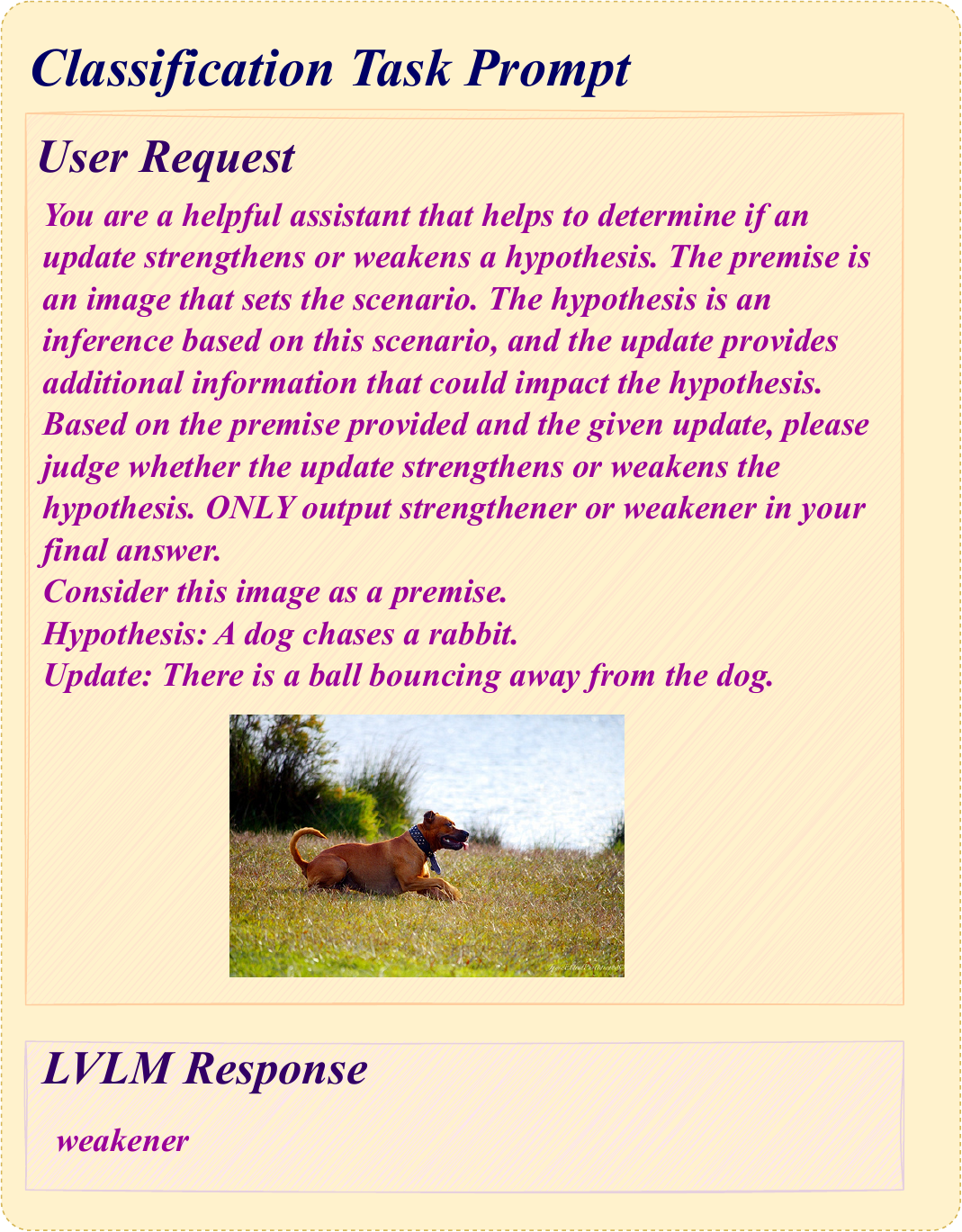}
    \caption{Prompt used for the Classification Task across LVLMs.}
    \label{fig:prompt_cls}
\end{figure}

We illustrate our prompt for the classification task for all the LVLMs in Figure \ref{fig:prompt_cls}.

\subsection{Prompt for Zero-shot Baselines in the Generation Task}
\begin{figure}[t]
    \centering
    \includegraphics[width=\linewidth]{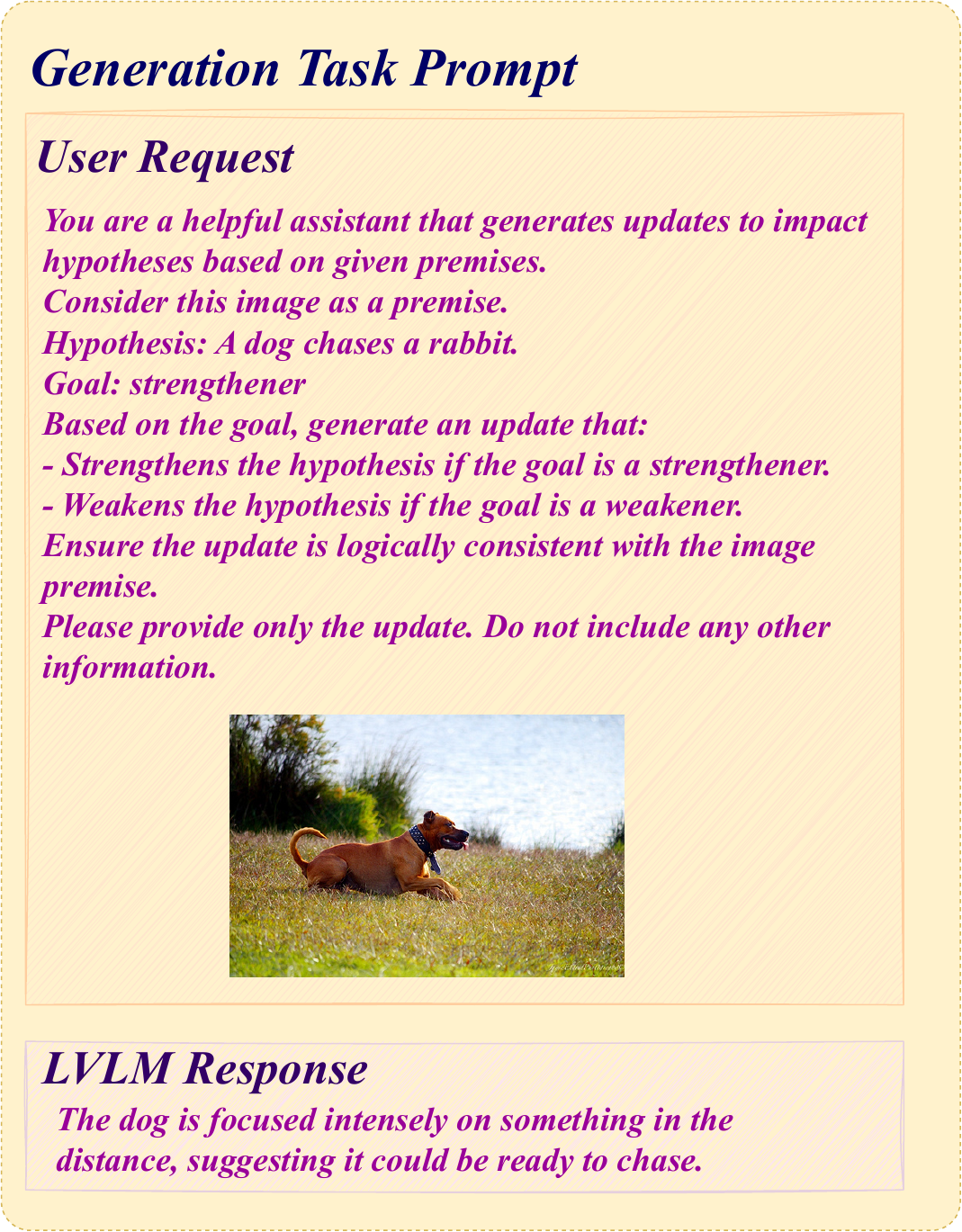}
    \caption{Prompt used for Zero-shot Baselines in the Generation Task.}
    \label{fig:prompt_gen}
\end{figure}

Our prompt for the generation task, applied consistently across all models, is shown in Figure \ref{fig:prompt_gen}.

\subsection{Prompt for Optimized Method}
\begin{figure*}[t]
    \centering
    \includegraphics[width=\linewidth]{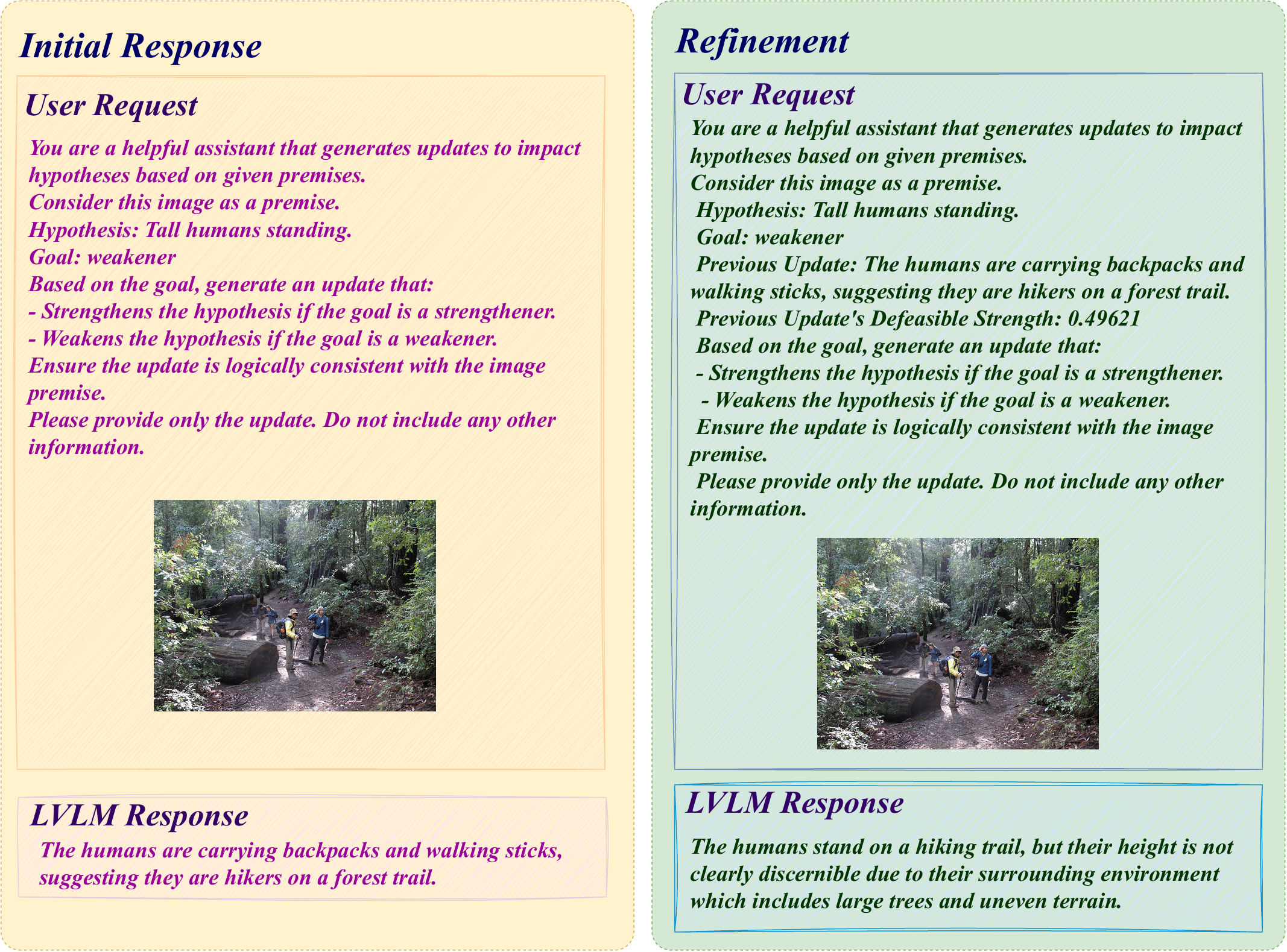}
    \caption{Prompt used for the Optimized Method.}
    \label{fig:prompt_opt}
\end{figure*}

We provide the prompt used for the optimized method, as illustrated in Figure \ref{fig:prompt_opt}.

% \section{B Experimental Details}

\section{Case Study for Evaluator}
\begin{figure}[t]
    \centering
    \includegraphics[width=\linewidth]{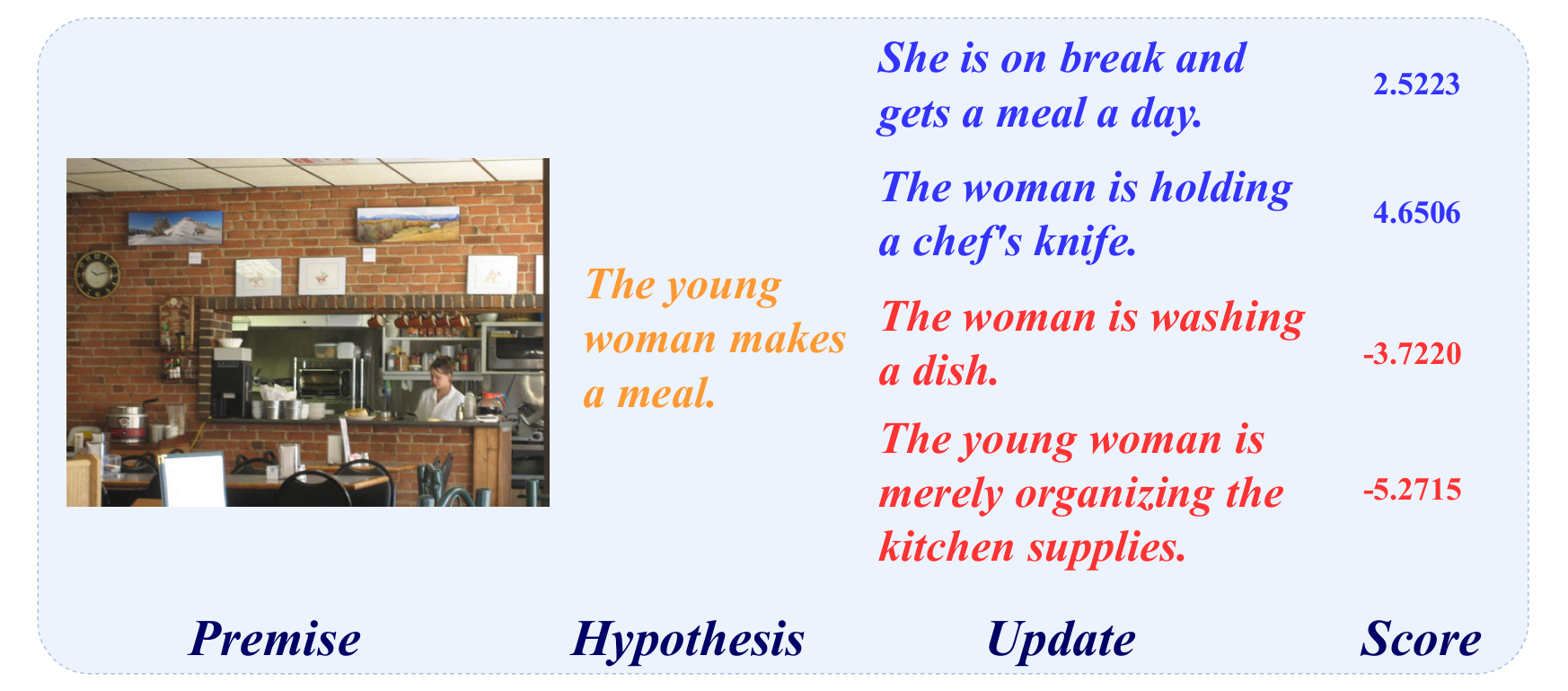}
    \caption{An example of evaluator scoring for different updates in visual entailment task. Blue indicates strengtheners and red indicates weakeners.}
    \label{fig:evaluator_case}
\end{figure}

To learn the qualitative performance of our metrics, we show cases in Figure \ref{fig:evaluator_case}.
We found our evaluator can accurately score updates to represent their entailment strengths. 
In particular, the update of strengthener  “She is on break and gets a meal a day” receives a moderate score of 2.5223. This update suggests she has time to eat but does not directly indicate she is cooking since the meal can be made by others, thus moderately supporting the hypothesis. In contrast, the update “The woman is holding a chef's knife” scores 4.6506, indicating a high level of support. A chef's knife is typically used in cooking, and given the kitchen setting, this update directly supports the hypothesis that she is making a meal. A similar observation can be found in the weakener example.
% For weakener, the update “The woman is washing a dish” scores -3.7220, weakening the hypothesis as it suggests she is cleaning up rather than cooking, though it still leaves the possibility that she was cooking beforehand. The update “The young woman is merely organizing the kitchen supplies” scores -5.2715, strongly contradicting the hypothesis by explicitly stating she is only organizing, thus excluding the possibility of her cooking.

\section{DVE and Related Datasets}
We further compare our DVE dataset with other related datasets, dividing them into two categories: visual understanding datasets and defeasible inference datasets. Table \ref{table:dataset_comparison} provides a detailed comparison. Most visual understanding datasets like SNLI-VE \cite{DBLP:journals/corr/abs-1901-06706}, VQA-v2.0 \cite{DBLP:conf/iccv/AntolALMBZP15}, and CLEVR \cite{DBLP:conf/cvpr/JohnsonHMFZG17} primarily focus on evaluating models' capabilities to interpret and reason about fixed, predefined visual scenes. However, they do not assess the models' ability to handle dynamic and nuanced semantic changes introduced by new, uncertain information. In contrast, DVE introduces the concept of defeasibility to tackle these uncertainties, thereby improving its capability to evaluate models' performance in reasoning with dynamic and uncertain information. Natural language inference datasets like $\delta$-NLI \cite{DBLP:conf/emnlp/RudingerSHBFBSC20} and $\delta$-CAUSAL \cite{DBLP:journals/corr/abs-2401-03183} introduce defeasibility in the entailment task but lack a metric to assess the impact of new information and overlook the defeasible inference in visual modality. In contrast, DVE incorporates visual information and provides an evaluator that reflects the strength of entailment brought by new information.
\begin{table}[ht]
    \centering
    \resizebox{0.45\textwidth}{!}{
    \begin{tabular}{lcccc}
        \toprule
        \textbf{Dataset} & \textbf{Multimodal} & \textbf{Strengthener} & \textbf{Weakener} & \textbf{\makecell{Entailment \\ Strength Metric}} \\
        \midrule
        \rowcolor{gray!10} \multicolumn{5}{c}{Visual understanding datasets}  \\
        \quad SNLI-VE & \cmark & \xmark & \xmark & \xmark \\
        \quad VQA-v2.0 & \cmark & \xmark & \xmark & \xmark \\
        \quad CLEVR & \cmark & \xmark & \xmark & \xmark \\
        \midrule
        \rowcolor{gray!10} \multicolumn{5}{c}{Natural language inference datasets}  \\
        \quad SNLI & \xmark & \xmark & \xmark & \xmark \\
        \quad $\delta$-NLI & \xmark & \cmark & \cmark & \xmark \\
        \quad $\delta$-CAUSAL & \xmark & \cmark & \cmark & \xmark \\
        \midrule
        \quad DVE & \cmark & \cmark & \cmark & \cmark \\
        \bottomrule
    \end{tabular}
    }
    \caption{Comparison of DVE and related datasets.}
    \label{table:dataset_comparison}
\end{table}

\section{Implementation Detail for Evaluator}
For the evaluation model, we select answers from LLaVA-1.5 \cite{DBLP:conf/nips/LiuLWL23a} and GPT-4o.
We employ a pre-trained BERT-large-uncased model \footnote{\url{https://huggingface.co/google-bert/bert-large-uncased}.} for text encoding and a ResNet50 model\footnote{\url{https://download.pytorch.org/models/resnet50-0676ba61.pth}.} for visual feature extraction. We utilized Adam optimizer~\cite{DBLP:journals/corr/KingmaB14} with a batch size of 32. 
% An adaptive learning rate scheduler reduces the learning rate if no improvement is observed on the validation set for a specified duration. 
The initial learning rate is set to $5 \times 10^{-6}$, with a weight decay of $1 \times 10^{-4}$. Training is conducted for up to 20 epochs. 
% Checkpoints are saved upon improvement in validation accuracy, with the final model checkpoint selected based on the highest validation accuracy. 
% The batch size for validation and testing is set to 64. 
The random seed in our code is set to 42. we set the hyper-parameters $d_1$ to 2048, $d_2$ to 1024, and $\alpha$ to 0.9.

\section{Experimental Setup for Evaluating Models on DVE Tasks}
For the classification task, we trained the models for up to 20 epochs with a random seed set to 42 and a batch size of 32. Different learning rates were used for different models: $1 \times 10^{-5}$ for VILT, $1 \times 10^{-3}$ for FLAVA, and $1 \times 10^{-5}$ for CLIP. 
For the zero-shot baselines, we used the same prompt across all models, as detailed in the Appendix. Similarly, for the generation task, a consistent prompt was used for all LVLMs, also specified in the Appendix. For our reward-driven update optimization method, we set $\eta$ to 1, and $M$ to 3. The prompt for this method is also included in the supplement material.

\section{Case Study for Reward-driven Update Optimization}

\begin{figure}[t]
    \centering
    \includegraphics[width=\linewidth]{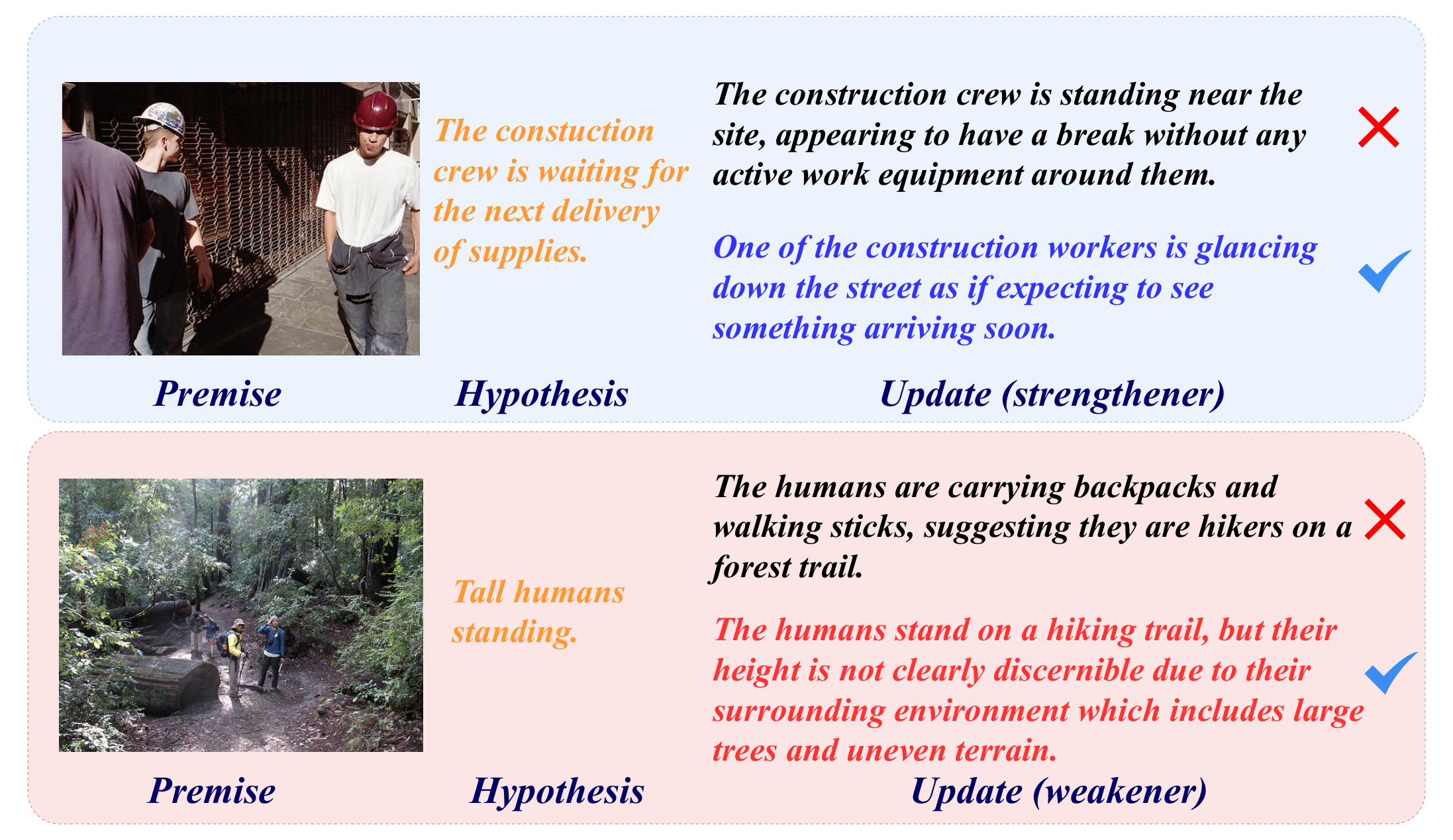}
    \caption{An example of the reward-driven update optimization method in action. The blue block indicates strengtheners and the red block indicates weakeners. The initial updates are in black, while the revised updates are in blue or red.}
    \label{fig:method_case}
\end{figure}

To further understand the performance of our proposed reward-driven update optimization method, we present an illustrative example in Figure \ref{fig:method_case}. In the strengthener case, the initially generated update posits that the crew is taking a break to enhance the hypothesis. However, suggesting that they are on a break may contradict the notion of waiting for supplies. The revised update, which describes a crew member glancing down the street as if expecting something to arrive soon, more effectively strengthens the hypothesis as it is closely related to the anticipation of the next delivery of supplies. A similar observation can be found in another weakener generation case.
% For the weakener case, the initial update attempts to weaken the hypothesis by emphasizing that the individuals are hikers, implying that they may be moving rather than standing still. However, this does not directly weaken the possibility that they may have been standing at some point, resulting in a relatively weak weakening effect. Conversely, the revised weakener effectively diminishes the likelihood of the hypothesis by stating that their heights are difficult to discern due to the surrounding environment, thus directly addressing the aspect of height.

\section{Effect from Writing Styles}
To verify that the evaluator is effective when it meets texts with a writing style that differs from the training dataset, we performed additional analyses using examples with varying styles generated by different models. 

The examples are as follows:

 \begin{itemize}
    \item The image premise is shown in Figure 1 of our paper
    \item \textbf{Hypothesis:} A dog chases a rabbit.
    \item \textbf{Strengtheners:}
    \begin{itemize}
        \item The dog looks like it’s going to chase something any second now. \textit{Score: 5.4776}
        \item Every muscle in the dog's body is alert, signaling it’s primed for a chase. \textit{Score: 5.5857}
        \item With that intense look, could the dog be any more ready to chase? \textit{Score: 5.4146}
    \end{itemize}
    \item \textbf{Weakeners:}
    \begin{itemize}
        \item A rabbit could photobomb this chase, and the dog would not even look up --- it’s got its eye on nothing else but its ball. \textit{Score: -4.3314}
        \item With a ball tossed by its owner, the dog's attention is fully absorbed in the game, showing zero interest in rabbits. \textit{Score: -4.3284}
        \item The dog is too absorbed in chasing the ball to even notice a rabbit. \textit{Score: -4.5062}
    \end{itemize}
\end{itemize}

Our results indicate that the evaluator consistently assigns comparable scores across these diverse updates, demonstrating its robustness to stylistic variations.

\section{Ablation for threshold and repetition}
We conducted this experiment using different thresholds and repetition numbers. The result is shown in Table \ref{tab:threshold_results}. 
\begin{table}[h!]
\centering
\begin{tabular}{lccc}
\hline
\textbf{Threshold} & \textbf{Round1} & \textbf{Round2} & \textbf{Round3} \\ \hline
$\pm0.5$ & 0.7956 & 0.8966 & 0.9384 \\
$\pm1.0$ & 0.8103 & 0.8916 & 0.9163 \\
$\pm1.5$ & 0.7438 & 0.8177 & 0.8670 \\
$\pm2.0$ & 0.6749 & 0.7562 & 0.8005 \\
\hline
\end{tabular}
\caption{Table showing results by threshold and round.}
\label{tab:threshold_results}
\end{table}

\section{Human Evaluation for Updates}

We conduct human evaluations for the generated updates, the results are in Table \ref{tab:human}

\begin{table}[ht]
\centering
\caption{Model Performance Comparison}
\begin{tabular}{lcc}
\toprule
\textbf{Model} &  & \\
  \textbf{Strengthener}   & \textbf{Ours} & \textbf{Human Annotation} \\
\midrule
InstructBLIP  & 0.7211 & 3.2081 \\
Multimodal-GPT      & 1.0690 & 3.1436 \\
MiniGPT-4           & 1.4998 & 3.6287 \\
mPLUG-Owl           & 2.2887 & 4.1084 \\
LLaVA-1.5           & 2.7868 & 4.1584 \\
GPT-4o              & 3.6413 & 4.6001 \\
GPT-4o (Optimized)  & 4.0679 & 4.6650 \\
\midrule
\textbf{Weakener}   & \textbf{Ours} & \textbf{Human Annotation} \\
\midrule
InstructBLIP        & 0.4231 & 3.1231 \\
Multimodal-GPT      & 0.7194 & 3.5226 \\
MiniGPT-4           & 0.9776 & 3.8276 \\
mPLUG-Owl           & 1.0274 & 3.7192 \\
LLaVA-1.5           & 0.4834 & 3.1773 \\
GPT-4o              & -2.5212 & 1.4680 \\
GPT-4o (Optimized)  & -2.9240 & 1.4335 \\
\bottomrule
\end{tabular}
\label{tab:human}
\end{table}

\end{document}